\documentclass[lettersize,journal]{IEEEtran}
\usepackage{amsmath,amsfonts}
\usepackage{algorithm}
\usepackage{array}
\usepackage[caption=false,font=normalsize,labelfont=sf,textfont=sf]{subfig}
\usepackage{textcomp}
\usepackage{stfloats}
\usepackage{url}
\usepackage{verbatim}
\usepackage{graphicx}
\usepackage{cite}

\usepackage[T1]{fontenc}
\usepackage{amssymb}
\usepackage{algorithm}
\usepackage{algorithmicx}
\usepackage{algpseudocode}
\usepackage{graphicx}
\usepackage{booktabs}
\usepackage{color}
\usepackage{comment}
\usepackage{multicol}

\floatname{algorithm}{Algorithm}

\usepackage{amsthm}
\newtheorem{remark}{Remark}
\usepackage{multirow}

\begin{document}

\title{Multi-Agent Deep Reinforcement Learning for Distributed and Autonomous Platoon Coordination via Speed-regulation over Large-scale Transportation Networks}
\author{Dixiao Wei, Peng Yi, Jinlong Lei, Xingyi Zhu
	\thanks{Dixiao Wei, Peng Yi, and Jinlong Lei are with  the Shanghai Research Institute for Intelligent Autonomous Systems, Tongji University, Shanghai, 201210, China, and Peng Yi and Jinlong Lei are also with the
		Department of Control Science and Engineering, Tongji University,  Shanghai, 201804, China;   Xingyi Zhu is with the Key Laboratory of Road and Traffic Engineering of the Ministry of Education, Tongji University, Shanghai, 201804, China.
        E-mails: {\tt\small \{weidx, yipeng,leijinlong\}@tongji.edu.cn}}
}


\maketitle

\begin{abstract}
Truck platooning technology enables a group of trucks to travel closely together, with which the platoon can save fuel, improve traffic flow efficiency, and improve safety.
In this paper, we consider the platoon coordination problem in a large-scale transportation network, to promote cooperation among trucks and optimize the overall efficiency. 
Involving the regulation of both speed and departure times at hubs, we formulate the coordination problem as a complicated dynamic stochastic integer programming under network and information constraints.
To get an autonomous, distributed, and robust platoon coordination policy, we formulate the problem into a model of the Decentralized-Partial Observable Markov Decision Process.
Then, we propose a Multi-Agent Deep Reinforcement Learning framework named Trcuk Attention-QMIX (TA-QMIX) to train an efficient online decision policy. 
TA-QMIX utilizes the attention mechanism to enhance the representation of truck fuel gains and delay times, and provides explicit truck cooperation information during the training process, promoting trucks' willingness to cooperate.
The training framework adopts centralized training and distributed execution, thus training a policy for trucks to make decisions online using only nearby information. Hence, the policy can be autonomously executed on a large-scale network. 
Finally, we perform comparison experiments and ablation experiments in the transportation network of the Yangtze River Delta region in China to verify the effectiveness of the proposed framework. In a repeated comparative experiment with 5,000 trucks, our method average saves 19.17\% of fuel with an average delay of only 9.57 minutes per truck and a decision time of 0.001 seconds.
\end{abstract}


\section{Introduction}

Platooning technology enables intelligent vehicles to travel closely together in a column in transportation networks while maintaining a fixed distance between each other \cite{bergenhem2012overview}.
This approach enhances transportation system efficiency and capacity, particularly beneficial in heavy-duty industries like truck transport \cite{robinson2010operating}. Notably, platooning can contribute to fuel efficiency and emission reduction. For example, \cite{tsugawa2014results} demonstrated a 13\% fuel consumption saving when three trucks formed a platoon with a 10-meter gap. Furthermore, if 40\% of trucks in the transportation network adopt platooning, it can result in a 2.1\% reduction in total carbon dioxide emissions.
Platooning technology has been developing rapidly \cite{aarts2016european,eilers2015companion,hewitt2005ensembles}. Particularly, platoon control technology has been extensively studied. Examples include the platoon safety issues in emergencies \cite{sidorenko2021safety}, platooning around curves \cite{wu2021curvilinear}, and the control of vehicle merging in platoons \cite{hu2020modelling}. Concurrently, vehicle-to-vehicle communication technology \cite{dey2016vehicle} and highway information architecture \cite{xu2014communication} have also garnered widespread attention. With the ongoing development of platoon control \cite{chu2019model} and communication technologies \cite{wu2022intelligent}, there is a promising prospect of forming a large-scale truck platoon traffic network in the future.

Platooning coordination studies how a group of vehicles forms a platoon, to maximize the benefits of platooning within various task and traffic network constraints, and it remains an active and challenging research topic \cite{lesch2021overview}.
\cite{larsen2019hub} focused on planning the departure time at hubs to wait for other trucks to form a platoon, and formulated the coordinated as an integer optimization problem to save fuel and enhance traffic efficiency simultaneously.
In \cite{johansson2018multi}, the authors explored platoon coordination for trucks sharing a common starting point but diverging at different endpoints from the departure hub. They formulated the problem as a non-cooperative game, wherein each truck seeks to maximize individual benefits within the platoon.
Considering the computational load of large-scale transportation networks, \cite{bai2021event} introduced distributed model predictive control to regulate truck waiting times at hubs. Additionally, the authors presented a distributed approximate dynamic programming scheme to address the coordination by waiting at hubs, ensuring compliance with hours-of-service regulations \cite{bai2022approximate}.
Moreover, \cite{johansson2022platoon,johansson2021real,bai2023large} studied the problem of coordination for multi-fleet platoons. \cite{johansson2021strategic} considered coordination under uncertain travel times but requires knowledge of the distribution of travel times in advance. 
All these works assumed that the travel time between two hubs is fixed and the coordination is fulfilled by planning the waiting time at hubs. This is not practical for a practical traffic network. 

On the other hand, path selection \cite{van2015fuel} and speed regulation \cite{abdolmaleki2021itinerary} can also be used for platoon coordination.
\cite{larsson2015vehicle} introduced a heuristic algorithm for pre-planning the path scheduling problem with
platooning coordination level as an optimization objective. \cite{liang2015heavy} explored a pairing algorithm for regulating speeds to form platoons.
\cite{van2017efficient,hoef2019predictive} discussed the combination of truck speed regulating and path scheduling for platoon coordination in a centralized planner. \cite{larson2014distributed} introduced a distributed and pairing coordination approach to determine optimal speed and route configurations. 
\cite{xiong2023approximate} designed junction-level coordination to plan the speed and routes of trucks at nearby hubs.
Nevertheless, the above methods' computation time tends to rise proportionally with the number of trucks within the network, limiting the online implementation in large-scale transportation networks, within which it is necessary for adapting to dynamic traffic scenarios.



Trucks often have different starting conditions, driving paths, and driving speeds.
This makes the traffic an ever-changing dynamic environment and requires adaptive platoon coordination.
The above existing methods can be used for re-planning once the environment changes, but they require whole-network information collection and solving complicated optimization problems, hindering the online implementation in large-scale networks.
Motivated by the above,  
we focus on autonomous coordination by simultaneously regulating speed and departure times, aiming for distributed and online decision-making to reduce both fuel consumption and time delays in a large-scale transportation network. 
We first formulate a distributed communication protocol for trucks, such that each truck makes decisions relying on local information in a rolling time domain, without the need for centralized information collection and coordination at a system level.
Then, we formulate the coordination problem as an integer optimization under dynamical stochastic constraints with decision variables of speed regulations on roads and hubs. Since  the  optimization problem is computationally intractable,
leveraging  Multi-agent Deep Reinforcement Learning (MADRL), we reformulate the problem as a Decentralized Partially Observable Markov Decision Process (Dec-POMDP). 
We propose an algorithm called Truck Attention-QMIX (TA-QMIX) within the Centralized Training and Distributed Execution (CTDE) framework. The policy network is designed to efficiently extract the spatial feature and temporal constraint of the transportation situation. And the potential coordination is also explicitly encoded in the policy network for improving training efficiency.
With the trained collaboration policy, the trucks can dynamically and rapidly regulate their speeds and departure times during their journeys in stochastic, high-dimensional, and dynamic traffic environments.
Moreover, each truck can leverage the well-trained policy to autonomous decisions with local traffic information.
We note that our earlier work \cite{wei2023multi} considered training a distributed policy for coordination at hubs by waiting within the MADRL framework.
Nevertheless, since speed regulation is additionally considered as a coordination approach, both the coordination problem formulation and its Dec-POMDP are reformulated, and a novel training algorithm called the TA-QMIX framework is proposed for improving training efficiency.

The main contributions are summarized as follows.

\begin{itemize}
	\item We design a novel platoon coordination problem to minimize fuel consumption and time delays by introducing state transition constraints with random noise and flexible cooperative approaches, unifying hub waiting and road rendezvous through speed regulation.
	
     \item We transform the complicated dynamical decision-making problem as a Dec-POMDP by designing distributed communication rules and dense reward functions.
     Under the distributed communication rules, each truck can make decisions based on local information. The complex platoon coordination problem can be efficiently solved through the Dec-POMDP model with dense rewards.
	
	\item We propose a novel MADRL algorithm called TA-QMIX within the CTDE framework for the formulated Dec-POMDP. 
 TA-QMIX combines the cross-attention mechanism and self-attention mechanism to promote training efficiency.
 The cross-attention block leverages cross-attention mechanisms to model inter-truck connections, and allows the truck to pay attention to location information and time information separately.
 The self-attention block concentrates on the truck's Q-value correlations and explicitly enhances the cooperative information flow, improving the total reward estimation.
 
	\item We set up simulation experiments in the Yangtze River Delta transportation network in China. Experiments verify the effectiveness and generalization of the TA-QMIX method. In a repeated comparative experiment involving 5,000 trucks, the proposed approach achieved an average of 19.17\% fuel savings with an average delay of only 9.57 minutes per truck. 
\end{itemize}

\begin{figure*}[t]
	\centering
	\includegraphics[scale=0.55]{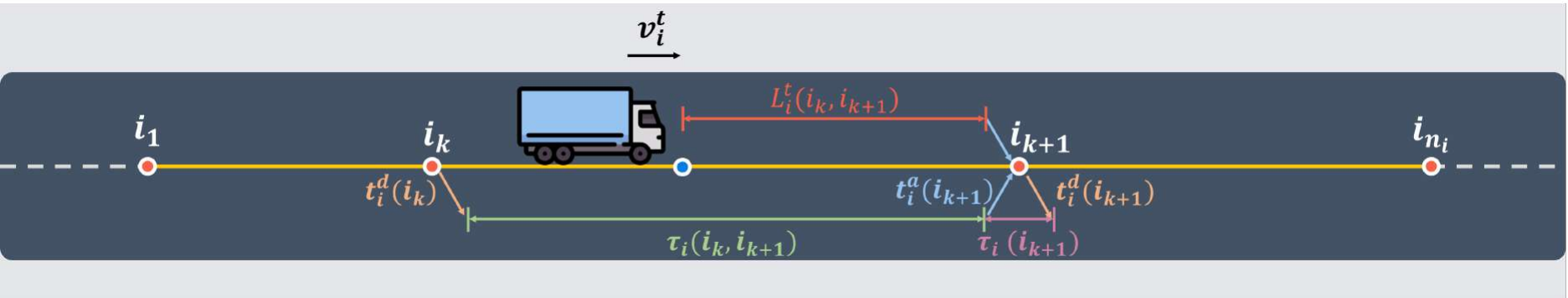}
	\caption{Truck $i$ traveling in the transportation network.}
	\label{fig1}
\end{figure*}

The rest of the paper is organized as follows.
Section II gives the considered platoon coordination optimization formulation, and its Dec-POMDP transformation with a complete scenario analysis. 
Section III introduces the TA-QMIX within the CTDE framework.
Section IV shows the ablation experiments and comparison experiments in a Chinese large-scale transportation network. 
Section V concludes this paper and outlines future research directions.
The main notations that are used in this paper are summarized in Table I.
\begin{table}[htb]   
	\begin{center}   
		\caption{Notations and explanations.}  
		\label{table:1} 
		\begin{tabular}{|c|c|}   
			\hline   \textbf{Notations} & \textbf{Explanations} \\
			\hline   $T_e$ & maximum time step \\  
			\hline   $\mathcal{N}$ &  the set of traveling trucks at time $t$\\ 
			\hline   $\mathcal{V}_i$ &   the hub set of truck $i$ routes \\  
			\hline   $\mathcal{E}_i$ &  the edge set of truck $i$ routes  \\     
			\hline   $\tau_i(i_k,i_{k+1})$ &   the travel time of truck $i$ on edge $(i_k,i_{k+1})$  \\ 
			\hline   $L(i_k,i_{k+1})$ &   the distance of edge $(i_k,i_{k+1})$  \\ 
            \hline   $S^t$	& the total state set at time $t$ \\ 
            \hline   $v^t$	& the total speed set at time $t$ \\ 
			\hline   $s_i^t$ &   the state of truck $i$ at time $t$ \\  
			\hline   $\mathcal{K}_d$ &   maximum tolerated delay time   \\ 
            \hline   $\Omega_i(S^t,v^t)$	& the set of trucks that co-traveling with truck $i$ \\ 
			\hline   $\phi_i(S^t,v^t)$ &   the drag coefficient of truck $i$ at time $t$\\ 
			\hline   $t_i^{a^\prime}(i_{n_i})$ & original arrival time of truck $i$\\ 
			\hline   $t_i^{a}(i_{n_i})$ & actual arrival time of truck $i$\\  
			\hline   $h_i(s_i^t)$ & the communication hub of truck $i$ at time $t$ \\  
			\hline   $m_i(s_i^t)$	& message sent by truck $i$ to the hub at time $t$\\ 
			\hline   $P_i(s_i^t)$	& the potential partner set of truck $i$  at time $t$\\ 
			\hline   $O_i(s_i^t)$	& observation function of truck $i$ \\ 
			\hline   $Z^t$	& observation set at time $t$ \\ 
            \hline   $F_r$	& fuel Consumption rate  \\ 
            \hline   $T_d$	& time Delay rate \\ 
            \hline   $P_j$	& platoon Journey rate \\ 
			\hline   $F_v$	& function value \\ 
			\hline   $T_o$	& timeout rate \\ 
            \hline   $P_r$	& participation rate \\ 
			\hline   
		\end{tabular}   
	\end{center}   
\end{table}

\section{Problem Formulation}
This section introduces platoon formation and the total platoon coordination optimization problem.

\subsection{Transportation Network and Platoon Formation} \label{sec2A}

We consider the platoon coordination within a transportation network $(V, E)$. Here, $V$ represents the set of hubs where trucks can wait to form platoons with other trucks, while $E$ represents the set of edges that connect the hubs in $V$. 
We consider a discrete-time setting and define $t \in \{1,2,\ldots, T_e\}$ with a time step $\Delta t$, where $T_e$ represents a maximal length allowing the last truck to reach its ending point. 
The set of trucks in the transportation network is represented by $\mathcal{N}=\{1,2,\ldots, N\}$.

Each truck $i\in\mathcal{N}$ has a starting point, an endpoint, and a pre-scheduled departure time $d_i \in \{1,2,\ldots, T_e\}$ from the starting point. It has a fixed path $\mathcal{E}_i = \{(i_1,i_2),(i_2,i_3),\ldots,(i_{n_i-1},i_{n_i})\}\subseteq E$  from the starting point $i_1$  to the endpoint $i_{n_i}$ by passing through hubs $\mathcal{V}_i=\{i_1,i_2,\ldots, i_{n_i}\}\subseteq V$.  
We denote $i_k$ as the $k$-th hub on the journey of truck $i$ for $k\in \{1,2,\ldots,n_i\}$. Meanwhile, $(i_k,i_k+1)$ represents the directed edge from hub $i_k$ to the next hub $i_{k+1}$ for $k\in \{1,2,\ldots,n_i-1\}$, with $L(i_k,i_{k+1})$ representing its length. The starting point $i_1$ is indexed by the 1-th hub while the ending point $i_{n_i}$ is indexed by the $n_i$-th hub.

During the journey, truck $i$ can adjust its speed  $v_i^t$ at time $t$ within a predefined discrete set  $ \{v_l,v_m,v_h,0\}$, such that it can accelerate or moderate on edges or wait at a hub to form a platoon with other trucks. This implies that truck $i$ travels at a constant speed $v_i^t$ within time interval $(t,t+\Delta t]$.
Here, $\{v_l,v_m,v_h\}$ denotes the low, medium, and high gear of speed that the truck can travel on edges. In addition, when truck $i$ is at a hub, $v_i^t$ can take $0$ representing that it stops and waits at the hub for an interval $(t,t+\Delta t]$. We assume that the length of each edge in the transportation network satisfies $L(i,j) > v_h  \Delta t$ for all $(i,j)\in E$.

To describe the system state, we define $s_i^t$ as a two-tuple  $(s_{i,1}^t,s_{i,2}^t)$ for truck $i$ at time $t$ as follows,
\begin{itemize}
	\item [(1)] $s_{i,1}^t$ represents the truck $i$'s location in its path $\mathcal{E}_i$. When truck $i$ is at hub $i_k,~k \in \{1,2,\ldots,n_{i}\}$ at time $t$, we denote by $s_{i,1}^t=i_k$. When truck $i$ is traveling on edge $(i_k, i_{k+1})$, $s_{i,1}^t=(i_k, i_{k+1})$.	
	\item [(2)] $s_{i,2}^t$ represents the distance of truck $i$ from the next hub, i.e., denote $s_{i,2}^t=L_i^t(i_k, i_{k+1})$ when  $s_{i,1}^t=(i_k, i_{k+1})$, and  $s_{i,2}^t=L(i_k, i_{k+1})$ when  $s_{i,1}^t=i_k$.
\end{itemize}
We denote the state of the transportation network at time $t$ as $S^t=\{s_1^t,s_2^t,\ldots,s_{N}^t\}$.

For brevity, we assume the truck is a point and the trucks form a platoon when they are at the same position with the same traveling speed. 
When truck $i$ is at the hub $i_k$, $k \in \{1,2,\ldots,n_{i}-1\}$, and there exists another truck $j$ at  the same hub $i_k$ with the same next hub, i.e., $k^\prime \in \{1,2,\ldots,n_{j}-1\}$, $s_i^t=s_j^t$, $i_{k+1}=j_{k^\prime+1}$,  truck $i$ and truck $j$ can form  a platoon on edge $(i_k, i_{k+1})$. 
If a group of trucks are at the same hub with the same next hub, they can form a platoon.

\begin{figure*}[t]
	\centering
	\subfloat[Catch up]{
        \includegraphics[scale=0.47]{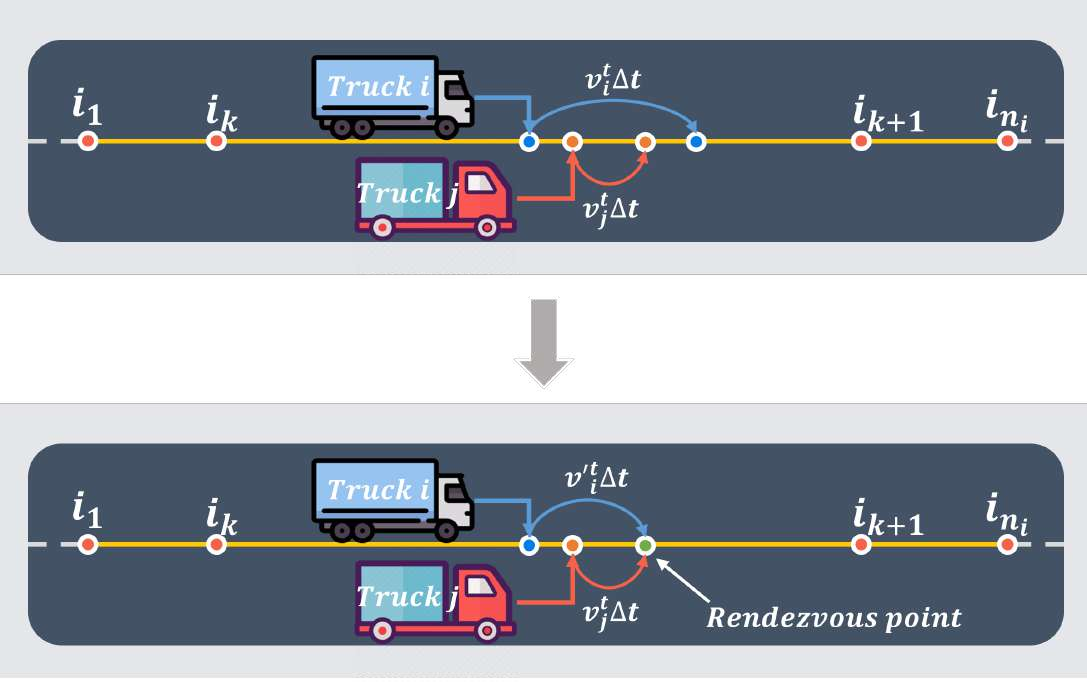}}
	\subfloat[Slow down]{
		\includegraphics[scale=0.47]{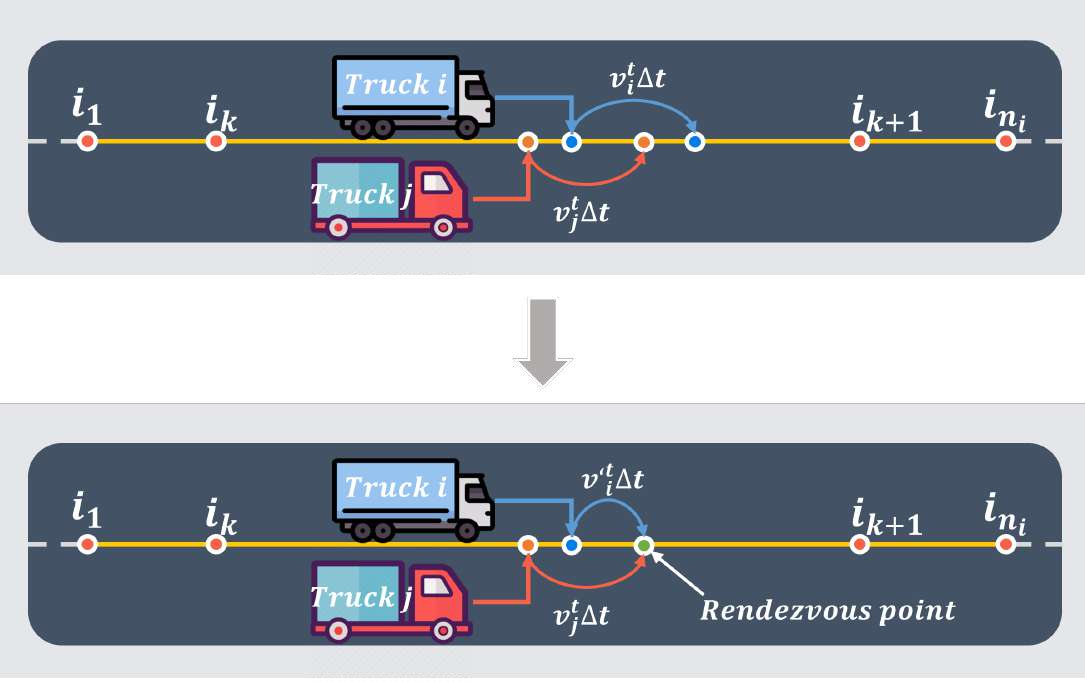}}
	\caption{Rendezvous  Processes. (a): truck $i$ is behind truck $j$ and catch up truck $j$. (b): truck $i$ is in front of truck $j$ and slow down for truck $j$.}
	\label{figalign}
\end{figure*}

When truck $i$ is traveling on the edge $(i_k, i_{k+1})$ and there exists another truck $j$ at the same position, $s_i^t=s_j^t$, they can also form a platoon. If a group of trucks are at the edge $(i_k, i_{k+1})$ at the same position, they can form a platoon. Otherwise, truck $i$ will travel alone. 
But the trucks can hardly rendezvous precisely when they are traveling on the edge, since we set three gears of speeds. Hence, we define two \textbf{Rendezvous Process}  for the trucks to rendezvous to form a platoon when they are at specific states with proper speeds as shown in Fig. \ref{figalign}.

Specially, when truck $i$ is traveling on the edge $(i_k,i_{k+1})$, there exists another truck $j$ traveling near by truck $i$, $j\neq i$, $s_{i,1}^t = s_{j,1}^t$, $0 < |s_{i,2}^t - s_{j,2}^t| < (v_h-v_l) \Delta t$. They can form a platoon
by the following two rendezvous processes.
  
\begin{itemize}
    \item \textbf{Catch up:} if truck $j$ is in front of truck $i$ , $s_{i,2}^t > s_{j,2}^t > v_h \Delta t$, and satisfying $0 < s_{i,2}^t - s_{j,2}^t < v_i^t \Delta t - v_j^t \Delta t$ as in Fig. \ref{figalign}(a), they should rendezvous during $(t,t+\Delta t]$. For simplicity, we just set they are at the same position the next time to form a platoon, $ s_{i,2}^{t+1} = s_{j,2}^{t+1} = s_{j,2}^t - v_j^t \Delta t$. 
    \item  \textbf{Slow down:} if truck $j$ is  behind truck $i$ , $s_{j,2}^t > s_{i,2}^t > v_h \Delta t$, and satisfying $0 < v_j^t \Delta t - v_i^t \Delta t < s_{j,2}^t - s_{i,2}^t < v_j^t \Delta t - v_l \Delta t$ as in Fig. \ref{figalign}(b), they should also rendezvous during $(t,t+\Delta t]$. We let $ s_{i,2}^{t+1} = s_{j,2}^{t+1} = s_{j,2}^t - v_j^t \Delta t$ at the next time step, and then they will form a platoon.
\end{itemize}

\begin{remark}  
In the above two processes, the truck can adjust its speed at a microscope level for rendezvous. For example,  truck $i$ can change its speed $v_i^t$ into $(s_{i,2}^t - s_{i,2}^{t+1}) / \Delta t$ and catch up truck $j$ in Fig. \ref{figalign}(a). 
\end{remark}

Furthermore, for multi-trucks traveling adjacent to each other on the same edge such that they should rendezvous at the next time interval, we can also define a rendezvous process.
Denote a set $M^t_{(i_k, i_{k+1})}$ such that it satisfies the following two conditions (i)  $\{ i \in M^t_{(i_k, i_{k+1})}   |  s_{i,1}^t=(i_k, i_{k+1}) \}$. 
(ii) for any pair of $i, j \in M^t_{(i_k,i_{k+1})}$, $|s_{i,2}^t-s_{j,2}^t|\leq \min(v_h-v_m,v_m-v_l)\Delta t$. 
The trucks in $M^t_{(i_k, i_{k+1})}$ can rendezvous by delicately adjusting their speed. Denote $k^*\in M^t_{(i_k,i_{k+1})}= \arg\max_{j\in  M^t_{(i_k,i_{k+1})} } \{ s_{j,2}^t - v_j^t \Delta t \}$,
we set the the state of all trucks in $M^t_{(i_k,i_{k+1})}$ as  $s_{j,2}^{t+1} = s_{k^*,2}^t - v_{k^*}^t \Delta t, \forall j\in M^t_{(i_k,i_{k+1})}$.
After a rendezvous process, trucks can travel together from the same position and form a platoon by traveling at the same speed. 

However, due to velocity deviations and road condition variations, there should be noise during the state transition. 
We add Gaussian noise to the position after state transition when the trucks travel on the edges.
Then, we can define the state transition dynamics of truck $i$ under a given transportation network state $S^t$, a velocity profile of the whole network $v^t=\{v_1^t,\cdots, v_N^t\}$, and network state transition noise $\xi^t=\{\xi_1^t,\cdots, \xi_N^t\}$.
\begin{equation}
\label{transition}
s_i^{t+1} \! \! = \! \! g_i(S^t,v^t, \xi^t) \! \! \triangleq  \! \!
	\begin{cases}
		((i_k,i_{k+1}), L(i_k,i_{k+1}) - v^t_i \Delta t + \xi_i^t) \text{,} & \\
		\quad  \text{if }s_{i,1}^t=i_k,v^t_i \in \{v_l,v_m,v_h\}, \\
  
		(s_{i,1}^t, L(i_k,i_{k+1})) \text{,} & \\
		\quad  \text{if }s_{i,1}^t=i_k,v^t_i=0,\\

        (s_{i,1}^t, s_{k^*,2}^t - v_{k^*}^t \Delta t + \xi_{k^*}^t) \text{,} & \\
		\quad  \text{if }s_{i,1}^t=(i_k,i_{k+1}),s_{i,2}^t>v^t_i\Delta t - \xi_{k^*}^t, \\
        \quad \quad i \in M^t_{(i_k,i_{k+1})},\\
  
        (s_{i,1}^t, s_{i,2}^t -v^t_i \Delta t + \xi_i^t) \text{,} & \\
		\quad  \text{if }s_{i,1}^t=(i_k,i_{k+1}),s_{i,2}^t>v^t_i\Delta t - \xi_i^t,\\
        \quad \quad i \notin M^t_{(i_k,i_{k+1})},\\
		
		(i_{k+1}, L(i_{k+1},i_{k+2})) \text{,} & \\
		\quad  \text{if }s_{i,1}^t=(i_k,i_{k+1}),s_{i,2}^t\leq v^t_i \Delta t - \xi_i^t.
	\end{cases}
\end{equation}
The state transition of the whole network is the cascading of all trucks under the above dynamics. Notice that when a truck $i$ is within a 
platoon at time $t$, i.e., there exists another truck $j$ that has the same position and velocity during $(t,t+\Delta t]$,  they should have the same
state transition noise to maintain the platoon at the next time. In other words, the state transition noise is added as a whole for a platoon by treating the platoon as a single truck. This makes the network state transition noise $\xi^t$ a state-dependent random vector. 
The above is not explicitly expressed in  \eqref{transition} just for clarity.

\subsection{Platoon Coordination Problem}
We form the platoon coordination problem by defining its decision variable, feasible set, and objective functions. The trucks can coordinate by changing their speed to form a platoon. 
If a truck $i$ or a truck group is at the hub, it can decide to wait by setting its speed to $0$ for forming a platoon with incoming trucks.
Meanwhile, when the trucks are on the edge, they can decide to accelerate or moderate to rendezvous to form a platoon.
 But notice that the feasible action of changing speed is also constrained by its state as
\begin{equation}\label{eqaction}
		v_i^t\in A_i(s_i^t) \triangleq 
		\begin{cases}
            \{v_l,v_m,v_h,0\} \text{,} &  \text{if } s_{i,1}^t=i_k,\\
		\{v_l,v_m,v_h\} \text{,} & \text{if } s_{i,1}^t=(i_k,i_{k+1}). 

		\end{cases}
\end{equation}

The objective function is to maximize the saving of fuel consumption in the entire transportation network meanwhile to minimize the time delay for the completion of transportation tasks. 
Next follows how to evaluate the time delay and fuel consumption under a given decision.

\textbf{Time delay:}
As shown in Fig. \ref{fig1}, for truck $i\in\mathcal{N}$ is at hub $i_{k+1}$, its departure time from  hub $i_{k+1}$ is defined as

\begin{equation}
	\label{eq1}
	t_i^d(i_{k+1}) = t_i^a(i_{k+1}) + \tau_i(i_{k+1}),
\end{equation}
where $t_i^a(i_{k+1})$ denotes the arrival time of truck $i$ at hub $i_{k+1}$ and $\tau_i(i_{k+1})\in \mathbb{N}$ is the waiting time of truck $i$ at hub $i_{k+1}$. 
$\tau_i(i_{k+1}) = |\{v^t_i|v^t_i=0, s_{i,1}^t = i_{k+1}, t \in \{1,2,\ldots, T_e\}\}|$. In particular, $t_i^d(i_1) = d_i+ \tau_i(i_1)$. The arrival time of truck $i$ at hub $i_{k+1}$ is given by
\begin{equation}
	\label{eq2}
	t_i^a(i_{k+1}) = t_i^d(i_k) + \tau_i(i_k,i_{k+1}),
\end{equation}
where $\tau_i(i_k,i_{k+1})$ represents the travel time of truck $i$ on edge $(i_k,i_{k+1})$. 
Let $L(i_k,i_{k+1})$ denotes the length of edge $(i_k,i_{k+1})$ and $v_i^t$ the speed  of truck $i$ at time $t$.
We assume that $\Delta t$ is a small step and the truck $i$ travels at a uniform speed $v_i^t$ within $(t,t+\Delta t]$.
$\tau_i(i_k,i_{k+1})$ can be calculated by both the length $L(i_k,i_{k+1})$  and the speed sequence  of truck $i$ when traveling over $(i_k,i_{k+1})$.
\begin{equation}
    \label{tauedge}
    \tau_i(i_k,i_{k+1}) = \mathop{\min} \{t| \sum_{x  = t_i^d(i_k)}^{t} v_i^x \Delta t \geq L(i_k,i_{k+1})\} -  t_i^d(i_k).
\end{equation}

For each truck $i$, waiting at a hub or traveling at a slower speed will delay the time to arrive at the ending point $i_{n_i}$. 
We define $t_i^{a^\prime}(i_{n_i})$ as the normal arriving time for truck $i$ to arrive at the ending point $i_{n_i}$. Under normal circumstances, the truck does not wait at any hub and travels at a fixed medium speed $v_m$. And the normal arrival time $t_i^{a^\prime}(i_{n_i})$ can be defined as

\begin{equation}
	\label{tia'}
	t_i^{a^\prime}(i_{n_i}) = d_i + \sum_{k=1}^{n_{i-1}}{\lceil L(i_k,i_{k+1}) / v_m \rceil}.
\end{equation}
Due to different traveling strategies, the actual arriving time $t_i^a(i_{n_i})$ that truck $i$ reaches the ending point $i_{n_i}$ may change:
\begin{equation}
	\label{tia}
	t_i^a(i_{n_i}) = d_i + \sum_{k=1}^{n_i-1}{\tau_i(i_k,i_{k+1})+ \tau_i(i_k)}.
\end{equation}
It means that when $t = t_i^a(i_{n_i})$, truck $i$ arrives at its end point, i.e., $s_{i,1}^t = i_{n_i}$.

Considering an arrival time delay budget rate $\mathcal{K}_d$, we wish that $t_i^a(i_{n_i})-d_i \leq \mathcal{K}_d  (t_i^{a^\prime}(i_{n_i})-d_i)$.
Denote $S=\{S^1,\cdots,S^{T_e}\}$ and $v=\{v^1,\cdots,v^{T_e}\}$, we define a delay ratio function as the minimizing objective 
\begin{equation}
    \label{timerate}
	J_{i,d}(S,v) = 
	\begin{cases}
            \frac{t_i^{a^\prime}(i_{n_i}) - t_i^a(i_{n_i})}{t_{total}}, & \text{if }  \frac{t_i^a(i_{n_i}) - d_i}{t_i^{a^\prime}(i_{n_i}) - d_i} \leq \mathcal{K}_d,\\
            c_b\frac{t_i^{a^\prime}(i_{n_i}) - t_i^a(i_{n_i})}{t_{total}}, & \text{otherwise}.\\
	\end{cases}
\end{equation}
Here, $t_{total} = \sum_{i \in \mathcal{N}}t_i^{a^\prime}(i_{n_i}) - d_i$ represents the sum of the normal traveling times of all trucks. 

\textbf{Fuel consumption:} The fuel consumption model of the truck platoon coordination problem is analyzed in detail in \cite{liang2015heavy}, which omits the fuel consumption due to acceleration and deceleration, friction, and gravity during the driving journey. Air drag resistance is the core measure of fuel consumption and can be defined as

\begin{equation}
	J_{air} = c_e \int \frac {1}{2} \rho_{a} \mathcal{A}_s \phi v^2 \, dl,
\end{equation}
where $c_e$ denotes the fuel energy coefficient, $\rho_a$ is the air density, $\mathcal{A}_s$ represents the truck's cross-sectional area, $\phi$ is the air drag coefficient, $v$ the velocity of truck, $l$ the traveling distance.
Thus, at time $(t,t+\Delta t]$, the fuel consumption of truck $i$ is
\begin{equation}
	\label{aircost}
	J_{i,a}(S^t,v^t) = \frac {1}{2} c_e \rho_a \mathcal{A}_s \phi(S^t,v^t) (v^t_i)^2 v^t_i \Delta t.
\end{equation}
Here, $\phi_i(S^t,v^t)$ represents the comprehensive drag coefficient related to whether truck $i$ travels in a platoon.
Extensive experiments have shown that the fuel savings for followers in a platoon are roughly the same, and leaders almost do not save fuel \cite{davila2013environmental}\cite{bishop2017evaluation}.
Similar to \cite{johansson2021strategic}, we distribute the fuel benefits obtained by the platoon equally among each truck in the platoon.
Let $\Omega_i(S^t,v^t) = \{j| j \in \mathcal{N}, j \neq i,s_j^t=s_i^t, v^t_j = v^t_i\}$ represents a set of trucks that travel in the same platoon with truck $i$.
Then, we have 
\begin{equation}
	\label{phi}
	\phi_i(S^t,v^t) = 
	\begin{cases}
		1 - (\phi_s - \phi_p) \frac{|\Omega_i(S^t,v^t)|}{|\Omega_i(S^t,v^t)| + 1}, & \text{if } \Omega_i(S^t,v^t) \neq \varnothing,\\
		\phi_s, & \text{otherwise}.
	\end{cases}
\end{equation}
Here, $\phi_s$ represents the air drag coefficient of the truck while alone, and $\phi_p$ represents the air drag coefficient of the truck when traveling in a platoon.
From Eq. (\ref{tia'}) and Eq. (\ref{aircost}), we have the normal fuel consumption of truck $i$ in the entire journey:
\begin{equation}
    J_{normal}(i) = \frac {1}{2} c_e \rho_a \mathcal{A}_s \phi_s (v_m)^2 v_m (t_i^{a^\prime}(i_{n_i}) - d_i).
\end{equation}
Thus,  at time $(t,t+\Delta t]$, we can define the fuel saving ratio of truck $i$ as follows:
\begin{equation}
    \label{fuelrate}
    \begin{aligned}
        J_{i,f}(S^t,v^t) & = \frac{\frac {1}{2} c_e \rho_a \mathcal{A}_s \phi_s (v_m)^2 v_i^t \Delta t - J_{i,a}(S^t,v^t)}{\sum_{i \in \mathcal{N}} J_{normal}(i)} \\
        & = \frac{(\phi_s (v_m)^2 - \phi_i(S^t,v^t) (v_i^t)^2) v_i^t \Delta t} {(v_m)^3 t_{total}}.
    \end{aligned}
\end{equation}

In summary, the whole platoon coordination problem can be formulated as
\begin{equation}
	\label{IO_formulation}
	\begin{array}{ll}
  		\max\limits_{v_i^t} & \sum_{i\in \mathcal{N}} \sum_{t=1}^{T_e} w_{1} {J_{i,f}(S^t,v^t)} +  w_{2} J_{i,d}(S,v)  \\
		\text { subject to } & \\

		& v_i^t \in A_i(s_i^t),    \forall i \in \mathcal{N}\\
        & s_i^{t+1} = g_i(S^t,v^t,\xi^t), \forall i \in \mathcal{N} 
	\end{array}
\end{equation}

Now the platoon coordination problem is formulated as a stochastic integer optimization with discrete decision variables and complicated hybrid stochastic dynamics constraints.
The following critical thinking motivates us to seek an adaptive and distributed online coordination policy with MADRL.

This problem \eqref{IO_formulation} is challenging to solve due to its huge decision space and hybrid dynamics constraints. 
The decision space is related to the number of trucks and the length of the coordination time horizon and is exponential at the scale of $3^{\mathcal{N}T_e}$. 
The state transition noise is state-dependent, hence decision-dependent and non-IID random vector. 
On the other hand, even if it can be solved before the start by collecting all the information, centralized optimization-based coordination has the following disadvantages.
1) Collecting full information of all trucks can be unaffordable, and even impossible for large-scale coordination problems due to privacy or conflict of interest. 
Moreover, each truck must accept regulations from the central node and cannot autonomously make decisions.
2) The method lacks adaptability and robustness. The schedule must be recomputed once the traffic conditions change or an unforeseeable accident happens.
And each time a new truck joins the transportation network, the entire strategy needs to be re-planned.
Thereby, the scheduling strategy can be only used for closed systems. 

We aim to provide a distributed coordination policy with the following property.
1) Each truck can online and autonomously make decisions with neighboring truck states and local traffic information.
2) The policy can adapt to various traffic scenarios and time-varying traffic conditions.
3) The policy can scale to large transportation networks and large time horizons.

To achieve those goals, we resort to the MADRL method, which has been proven to be effective in solving sequential decision-making problems in dynamic environments\cite{huang2023multi}, which from our viewpoint is also the first try to adopt MADRL for speed regulation platoon coordination.

\section{Dec-POMDP Formulation for Platoon Coordination}\label{sec:state}
In this section, we reformulated this problem as a Dec-POMDP for training a distributed and adaptive policy with MADRL. The Dec-POMDP problem is composed of a tuple $G=(\mathcal{N},S,\mathcal{A},T,R,Z,O)$ \cite{oliehoek2016concise}. 
In the following, we give $G$ at every time $t$ in detail.

\textbf{State Space:} State space $S^t=\{s_1^t,s_2^t,\ldots, s_N^t\}$, where each $s_i^t, i\in \mathcal{N}$ is defined in \ref{sec2A}.

\textbf{Action Space:}
At time $t$, truck $i$ can take the action $a_i^t$ of changing its speed $v_i^t \in A(s_i^t) $ with the constraint Eq. (\ref{eqaction}).
Due to the noise $\xi$, it is often difficult for trucks to be in the same position. After trucks rendezvous in the same position, we assume that these trucks have achieved the intention to form a platoon, that is trucks in the same position can only travel at a speed of $v_m$ and form a platoon.
Let $\mathcal{X}_i(S^t) = \{j| j \in \mathcal{N}, j \neq i,s_j^t=s_i^t\}$ represents a set of trucks that are in the same platoon with truck $i$.
Thus, the action $a_i^t$ is constrained by
\begin{equation}
\label{eqa'}
    a_i^t \! \! \in \! \! A'_i(s_i^t) \! \! \triangleq \! \!
    \begin{cases}
    \{v_l,v_m,v_h,0\} \text{,} &  \! \! \! \! \! \text{if } s_{i,1}^t=i_k,\\
    \{v_l,v_m,v_h\} \text{,} &  \! \! \! \! \! \text{if } s_{i,1}^t=(i_k,i_{k+1}) , \mathcal{X}_i(S^t) = \varnothing, \\ 
    \{v_m\} \text{,} &  \! \! \! \! \! \text{if } s_{i,1}^t=(i_k,i_{k+1}) , \mathcal{X}_i(S^t)\neq \varnothing.
    \end{cases}
\end{equation}
The joint action space of all trucks is defined as $\mathcal{A}^t = v^t = \{a_1^t,a_2^t,\ldots,a_N^t\}$, $\mathcal{A} = v = \{\mathcal{A}^1,\mathcal{A}^2,\ldots , \mathcal{A}^{T_e}\}$. 

{\textbf{State Transition Function:} $T(S^t,\mathcal{A}^t,\xi^t): S^t\times \mathcal{A}^t\rightarrow S^{t+1}$ represents the function that $S^t$ transfers to $S^{t+1}$ after performing the joint action $\mathcal{A}^t$. The state transition function is defined in Eq. (\ref{transition}).

\textbf{Observation Space and Observation Function:}
The observation of a truck depends on the information it receives from other trucks. 
The trucks are assumed to have limited communication scope within large-scale transportation networks.
At time $t$, each truck $i \in \mathcal{N}$ can communicate with its neighboring trucks 
through a hub $h_i(s_i^t)$ defined as (\ref{eqh}).
\begin{equation}
	\label{eqh}
	h_i(s_i^t)=
	\begin{cases}
		i_{k} \text{,} & \text{if } s_{i,1}^t=i_k, \\
		i_{k+1} \text{,} &  \text{if } s_{i,1}^t=(i_k,i_{k+1}), \\
		
	\end{cases}
\end{equation}
This implies that each truck communicates with the trucks having the same next hubs.
For example, as shown in Fig. \ref{fignetwork}, the green truck $j$ located at hub $j_k^\prime$ will communicate through hub $i_k$, while the blue truck $i$   driving on edge $(i_{k},i_{k+1})$ will communicate through hub $i_{k+1}$. 
\begin{figure}[t]
	\centering
	\includegraphics[scale=0.29]{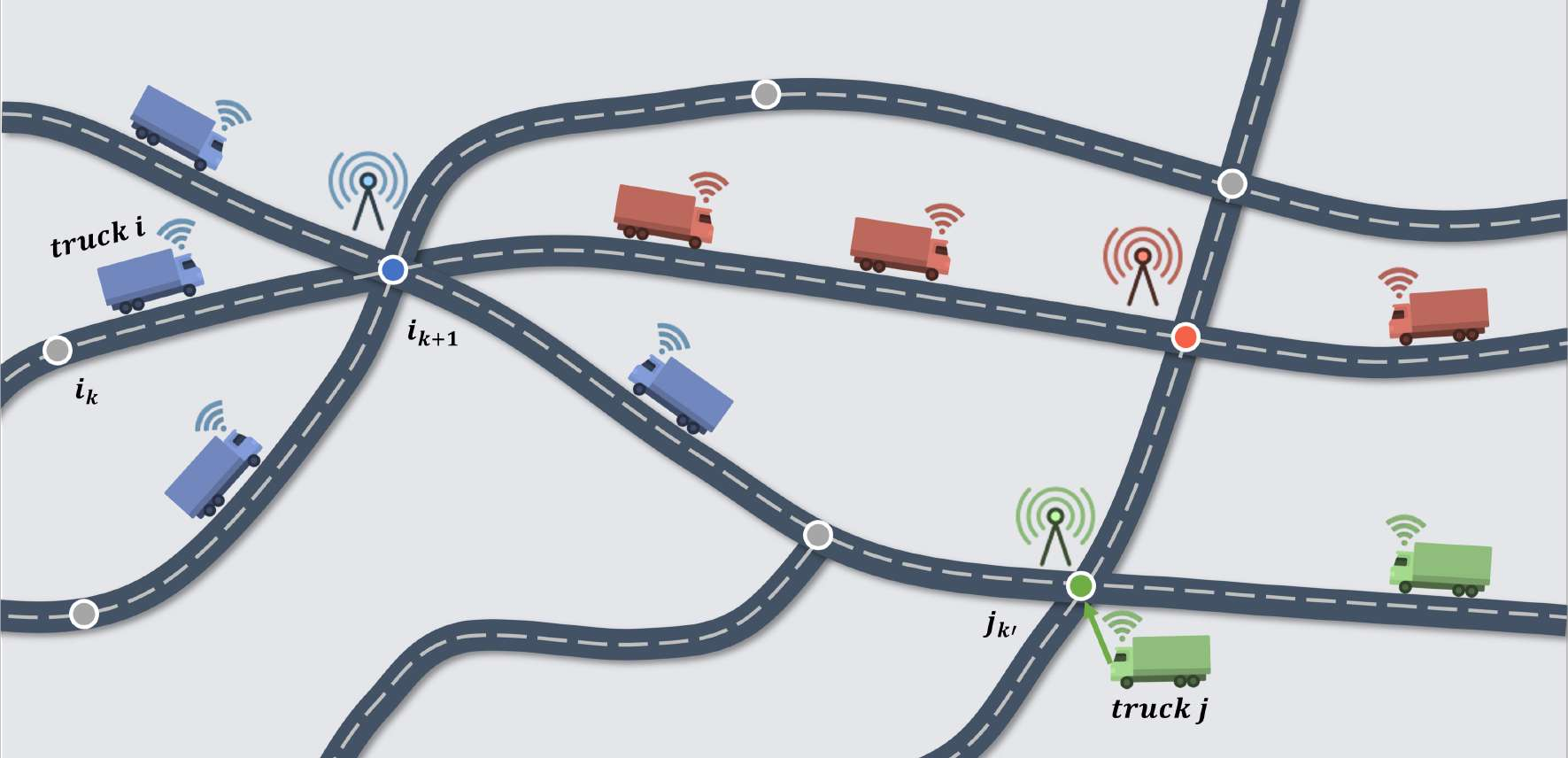}
	\caption{Distributed communication in transportation network. The trucks with the same color can communicate with each other. }
	\label{fignetwork}
\end{figure}

The communicated  information of truck $i$  through hub $h_i(s_i^t)$  is denoted by $m_i(s_i^t)=(m_{i,1}(s_i^t),m_{i,2}(s_i^t))$, where $m_{i,1}(s_i^t)$ denotes the next hub of the hub $h_i(s_i^t)$ that truck $i$ is going to, and  $m_{i,2}(s_i^t)=s_i^t$. Thus, $m_i(s_i^t)$ is defined by
\begin{equation}
	m_i(s_i^t) =
	\begin{cases}
		(i_{k+1},s_{i}^t) \text{,} &  \text{if } s_{i,1}^t=i_k, \\
		(i_{k+2},s_{i}^t) \text{,} &  \text{if } s_{i,1}^t=(i_k,i_{k+1}).
	\end{cases}
\end{equation}

The set of trucks that has the same communication hub $h_i(s_i^t)$ as truck $i$ is defined as
\begin{equation}
	\mathcal{C}_i(s_i^t)=\{j:h_j(s_j^t)=h_i(s_i^t),j \in \mathcal{N},j \neq i\}.
\end{equation}
Meanwhile, the potential partner set of truck $i$, representing  the set of trucks that may form a platoon with truck $i$, is denoted as
\begin{equation}
	\label{eqpartner}
	P_i(s_i^t) \!\!=\!\!
	\begin{cases}
		\{j:j \in \mathcal{C}_i(s_i^t), m_{i,1}(s_i^t)=m_{j,1}(s_j^t)\} \text{,} \\
		\quad \quad \quad \quad \quad \quad \quad \quad \quad \quad \quad \text{if } s_{i,1}^t=i_k, \\
		\{j:j \in \mathcal{C}_i(s_i^t), m_{i,1}(s_i^t)=m_{j,1}(s_j^t) \lor s_{i,1}^t=s_{j,1}^t\} \text{,} \\
		\quad \quad \quad \quad \quad \quad \quad \quad \quad \quad \quad \text{if } s_{i,1}^t=(i_k,i_{k+1}).
	\end{cases}
\end{equation}
When truck $i$ is at hub $i_k$, i.e., $s_{i,1}^t=i_k$, $P_i(s_i^t)$ means the trucks in the communication set $\mathcal{C}_i(s_i^t)$ that are also about to go to hub $i_{k+1}$. When truck $i$ is traveling on the edge $(i_k,i_{k+1})$, i.e., $s_{i,1}^t=(i_k,i_{k+1})$, $P_i(s_i^t)$ represents the trucks in the communication set $\mathcal{C}_i(s_i^t)$ that are also traveling on the edge $(i_k,i_{k+1})$ or about to go to hub $i_{k+2}$.

We can define truck $i$'s observation function to obtain information about itself and nearby trucks,  which can be described as
\begin{equation}
	\label{eqmr}
	z_i^t = O_i(s^t) = \{s_j^t:j \in P_i(s_i^t)\} \cup \{s_i^t\}.
\end{equation}
Specifically, when truck $i$ arrives at its destination $i_{n_i}$, it no longer communicates with hub $i_{n_i}$.
Observation space is denoted by $Z^t=\{z_1^t,z_2^t,\ldots,z_{|\mathcal{N}|}^t\}$.

\textbf{Reward:}
The goal of platoon coordination is to form as many platoons as possible to save fuel and reduce potential time delays while satisfying the task constraint of each truck.
According to Eq. (\ref{fuelrate}), during the whole journey, the truck gets a dense fuel reward  $w_{1} J_{i,f}(S^t,v^t)$ every time $t$.
When a truck arrives at its ending point, it gets a time delay reward $w_{2} J_{i,d}(S,v)$ from Eq. (\ref{timerate}).

Considering that sparse time rewards are not conducive to the training convergence of MADRL, we decompose the time objective function $J_{i,d}(S,v)$ into approximate dense time rewards $R_{i,d}(s_i^t,a_i^t)$ and an ending reward $R_{i,e}(S,\mathcal{A})$:
\begin{equation}
\label{eqtimedecompose}
    J_{i,d}(S,v) \approx \sum_{t=1}^{T_e} R_{i,d}(s_i^t,a_i^t) + R_{i,e}(S,\mathcal{A})
\end{equation}
When truck $i$ is traveling in the transportation network, it will obtain a dense time reward based on its action.
\begin{equation}
    \label{rewardt}
    R_{i,d}(s_i^t,a_i^t) = 
    \begin{cases}
        w_2 \frac{a_i^t \Delta t / v_m - \Delta t}{t_{total}} \text{,} & \text{if } s_{i,1}^t \neq i_{n_i}, \\
        0, & \text{otherwise}.
    \end{cases}
\end{equation}
When truck $i$ arrives at its ending point $i_{n_i}$ (i.e., $s_{i,1}^t=i_{n_i}$), it will obtain a ending reward $R_{i,e}(S,v)$:
\begin{equation}
    \label{rewarde}
    R_{i,e}(S,\mathcal{A}) \! \!   = \! \! \!
    \begin{cases}
        w_2 (c_b - 1)\frac{t_i^{a^\prime}(i_{n_i}) - t_i^a(i_{n_i})}{t_{total}} , & \! \! \text{if }  \frac{t_i^a(i_{n_i}) - d_i}{t_i^{a^\prime}(i_{n_i}) - d_i} > \mathcal{K}_d,\\
        0, & \! \! \text{otherwise}.
    \end{cases}
\end{equation}

According to \eqref{eqa'}, the truck would not change its speed after forming a platoon until it arrives at the next hub. We increase the reward when the truck forms a platoon to promote better training convergence. 
If truck $i$ forms a new platoon or join a platoon (i.e. truck $i$ is traveling alone at the last time $t-1$ and $|\Omega_i(S^t,\mathcal{A}^t)|$ is increasing at time $t$), it will receive a fuel reward based on the remaining distance $s_{i,2}^t$ and the air drag coefficient $\phi_i(S^t,\mathcal{A}^t)$.
If truck $i$ is traveling in a platoon and there are new trucks join the platoon (i.e. $\Omega_i(S^{t-1},\mathcal{A}^{t-1}) \neq \varnothing$ and $|\Omega_i(S^t,\mathcal{A}^t)| > |\Omega_i(S^{t-1},\mathcal{A}^{t-1})|$), it will also receive a fuel reward based on the remaining distance $s_{i,2}^t$ and the lower air drag coefficient $\phi_i(S^t,\mathcal{A}^t)$. Therefore, the overall reward design is given as follows.
\begin{equation}
    \label{rewardf}
    R_{i,f}(S^t,\mathcal{A}^t)  = 
    \begin{cases}
        w_1 J_{i,f}(S^t,\mathcal{A}^t),  \\
        \quad   \quad \text{if } \Omega_i(S^t,\mathcal{A}^t) = \varnothing,\\
        w_1 \frac{(\phi_s - \phi_i(S^t,\mathcal{A}^t)) s_{i,2}^t}{v_m t_{total}}, \\
        \quad   \quad \text{if } \Omega_i(S^t,\mathcal{A}^t) > |\Omega_i(S^{t-1},\mathcal{A}^{t-1})|, \\
        \quad \quad \quad \Omega_i(S^{t-1},\mathcal{A}^{t-1}) = \varnothing, \\
        w_1 \frac{(\phi_i(S^{t-1},\mathcal{A}^{t-1})) - \phi_i(S^t,\mathcal{A}^t))  s_{i,2}^t}{v_m t_{total}}, \\
        \quad   \quad \text{if } |\Omega_i(S^t,\mathcal{A}^t)| > |\Omega_i(S^{t-1},\mathcal{A}^{t-1})|, \\
        \quad \quad \quad \Omega_i(S^{t-1},\mathcal{A}^{t-1}) \neq \varnothing, \\
        0,  \quad  \text{otherwise}.
    \end{cases}
\end{equation}

We aim to find a policy for each truck as a mapping $ \rho_i: z_i^t\rightarrow a_i^t $ to maximize the following rewards:
\begin{equation}
\max \sum_{i \in \mathcal{N}} (\sum_{t=1}^{T_e}  R_{i,f}(S^t,\mathcal{A}^t) +  R_{i,d}(s_i^t,a_i^t)) + R_{i,e}(S,\mathcal{A}).
\end{equation}

\section{ TA-QMIX Algorithm for Training  Platoon Coordination policy}
This section elaborates on the operational process of training a platoon coordination policy by multi-agent reinforcement learning algorithms.

\subsection{Representation for State and Observation}
Neural networks usually require fixed-length of inputs. Since the dimensions of  $S^t=\{s_1^t,s_2^t,\ldots,s_{|\mathcal{N}|}^t\}$ and $Z^t=\{z_1^t,z_2^t,\ldots,z_{|\mathcal{N}|}^t\}$ are determined by the number of trucks, they can not be input directly to the neural network. Considering the variability of the transportation network scale, we need to design a representation of the state and observation into a fixed-size input.

The representation of the state is denoted as $\mathcal{S}^t=(\mathcal{S}_{edge}^t,\mathcal{S}_{co}^t)$.
At time $t$, $\mathcal{S}_{edge}^t$ represents the location information of trucks on the edge of the transportation network.
$\mathcal{S}_{co}^t$ represents the information about the successful cooperation of the truck.

First, for each edge $e_m\in E$, we discretize the length $L(e_m)$ into $\alpha$ segments as follows.
\begin{equation}
	\bar{L}(e_m) = (L(e_{m}[1]), L(e_{m }[2]), \ldots, L(e_{m}[\alpha])),
\end{equation}
where \begin{equation*}
	\begin{split}
		L(e_{m}[j]) = \left (\frac{\alpha-j}{\alpha}L(e_m),\frac{\alpha+1-j}{\alpha}L(e_m) \right],\\
		j \in \{1,2,\ldots,\alpha\}.
	\end{split}
\end{equation*}

We let the state of edge $e_m$ be the number of trucks driving on each segment of the edge at time $t$, i.e.,
\begin{equation}
	\mathcal{S}_m^t = \{|\mathcal{M}(e_{m}^t[1])|,|\mathcal{M}(e_{m}^t[2])|,\ldots,|\mathcal{M}(e_{m}^t[\alpha])|\}.
\end{equation}
Here, $\mathcal{M}(e_{m}^t[j]),j \in \{1,2,\ldots,\alpha\}$ denotes the set of trucks driving on the edge $e_m$, for which the remaining length belongs to the interval $L(e_{m}[j])$, i.e., $\mathcal{M}(e_{m}[j])=\{i:i \in \mathcal{N}, s_{i,1}^t=e_m, s_{i,2}^t \in L(e_{m}[j])\}$. 
Hence the state of all edges is denoted as $\mathcal{S}_{edge}^t = (\mathcal{S}_1^t, \mathcal{S}_2^t, \ldots, \mathcal{S}_{|E|}^t)$, which has a fixed size of $\alpha|E|$.

Second, $\mathcal{S}_{co}^t=(\mathcal{S}_{co}^t[1],\mathcal{S}_{co}^t[2],\dots,\mathcal{S}_{co}^t[\mathcal{N}])$ is a binary vector of length $|\mathcal{N}|$, where each element is either 0 or 1.
$|\mathcal{N}|$ is the maximum number of trucks carried in the transportation network. At time $t$, if truck $i$ forms a platoon, $\mathcal{S}_{co}^t[i] = 1$. Clearly, for $\forall i \in \{1,2,\dots,|\mathcal{N}|\}$, $\mathcal{S}_{co}^t[i]$ can be defined as
\begin{equation}
    \mathcal{S}_{co}^t[i] = 
	\begin{cases}
		1, & \text{if } \Omega_i(S^t,\mathcal{A}^t) \neq \varnothing ,\\
		0, & \text{otherwise}.\\
	\end{cases}
\end{equation}
Therefore, $\mathcal{S}^t$ has a fixed size of $\alpha|E| + |\mathcal{N}|$.

The observation $Z_i^t = O_i(s_i^t)$ changes with the number of potential partners that share information through hub $h_i(s_i^t)$, see (\ref{eqmr}). We define the representation of the observation as 
\begin{equation}
	\mathcal{Z}_i^t = \mathcal{O}_i(s_i^t) = (\mathcal{Z}_{i,self}^t,\mathcal{Z}_{i,delay}^t,\mathcal{Z}_{i,others}^t). 
\end{equation}
Here, $\mathcal{Z}_{i,self}^t = (s'^t_{i,1},s_{i,2}^t)$ represents truck $i$'s own location information. $s'^t_{i,1}$ divides the truck state $s_i^t$ into four categories: at a hub, traveling alone on an edge, traveling in a platoon, and inactive. Eq. (\ref{bars1}) shows the details.
\begin{equation}
	\label{bars1}
	s'^t_{i,1} =
	\begin{cases}
		0, & \text{if } s_{i,1}^t = i_k,\\
		1, & \text{if } s_{i,1}^t = (i_k,i_{k+1}), \mathcal{X}_i(S^t) = \varnothing,\\
		2, & \text{if } s_{i,1}^t = (i_k,i_{k+1}), \mathcal{X}_i(S^t) \neq \varnothing,\\
		3, & \text{otherwise}.
	\end{cases}
\end{equation}
$\mathcal{Z}_{i,delay}^t = \sum_{x=d_i}^{x=t} (a_i^x \Delta t / v_m - \Delta t) + \mathcal{K}_d (t_i^{a^\prime}(i_{n_i}) - d_i)$ represents the time budget of truck $i$.

$\mathcal{Z}_{i,others}^t$ represents information from $q$ trucks in the relative potential partner set ${RP}_i(s_i^t)$. 
Let ${RP}_i(s_i^t)=\{rs_j^t:j \in {P}_i(s_i^t)\}$. For $j \in {P}_i(s_i^t)$, $rs_j^t=(rs_{j,1},rs_{j,2})$. $rs_{j,1}$ and $rs_{j,2}$ represents the location information of truck $j$ relative to truck $i$ show as follow.
\begin{equation}
	rs_{j,1}^t = 
	\begin{cases}
		0, & \text{if } s_{j,1}^t = s_{i,1}^t,\\
		1, & \text{if } s_{j,1}^t \neq s_{i,1}^t.
	\end{cases}
\end{equation}
\begin{equation}
	rs_{j,2}^t = 
	\begin{cases}
		0, & \text{if } s_{j,1}^t = s_{i,1}^t,s_{j,1}=j_{k'}, s_{i,1}^t = i_k,\\
		-s_{j,2}^t, & \text{if } s_{j,1}^t = (j_{k'},j_{k'+1}),\\
		& \quad s_{i,1}^t = i_k,j_{k'+1}=i_k,\\
		s_{i,2}^t-s_{j,2}^t, & \text{if } s_{j,1}^t=(j_{k'},j_{k'+1}),s_{i,1}^t=(i_k,i_{k+1}),\\ 
		& \quad j_{k'+1}=i_{k+1}.\\
	\end{cases}
\end{equation}

By sorting the relative potential partner set ${RP}_i(s_i^t)$ in descending order according to $|rs_{j,2}^t|$ to obtain the sorted set $\mathcal{P}_i(s_i^t)$, $\mathcal{Z}_{i,others}^t$ is defined as the top $q$ elements from $\mathcal{P}_i(s_i^t)$.

Specially, if $|\mathcal{P}_i(s_i^t)| < q$, $\mathcal{P}_i(s_i^t)$ will be padded with $(3,0)$ to length $q$. Hence, for each truck $i$, its new observation $\mathcal{Z}_i^t$ has a fixed size of $2q+3$.

\begin{figure*}[t]
	\centering
	\includegraphics[scale=0.42]{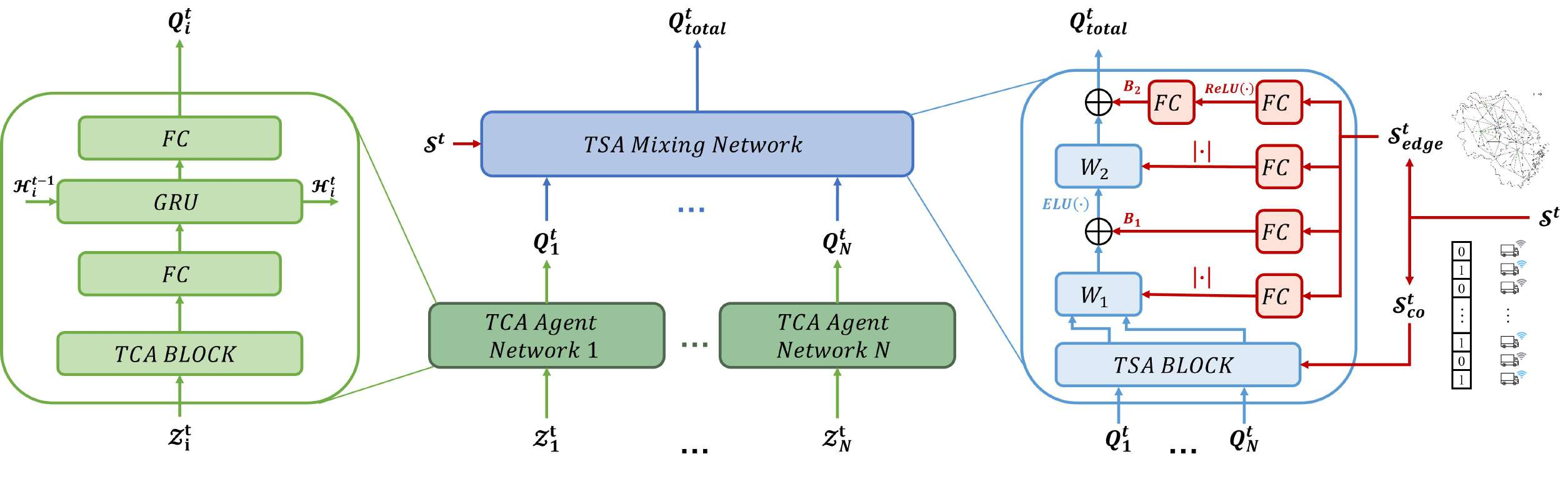}
	\caption{TA-QMIX structure.}
	\label{figtaqmix}
\end{figure*}

\begin{figure}[t]
	\centering
	\includegraphics[scale=0.27]{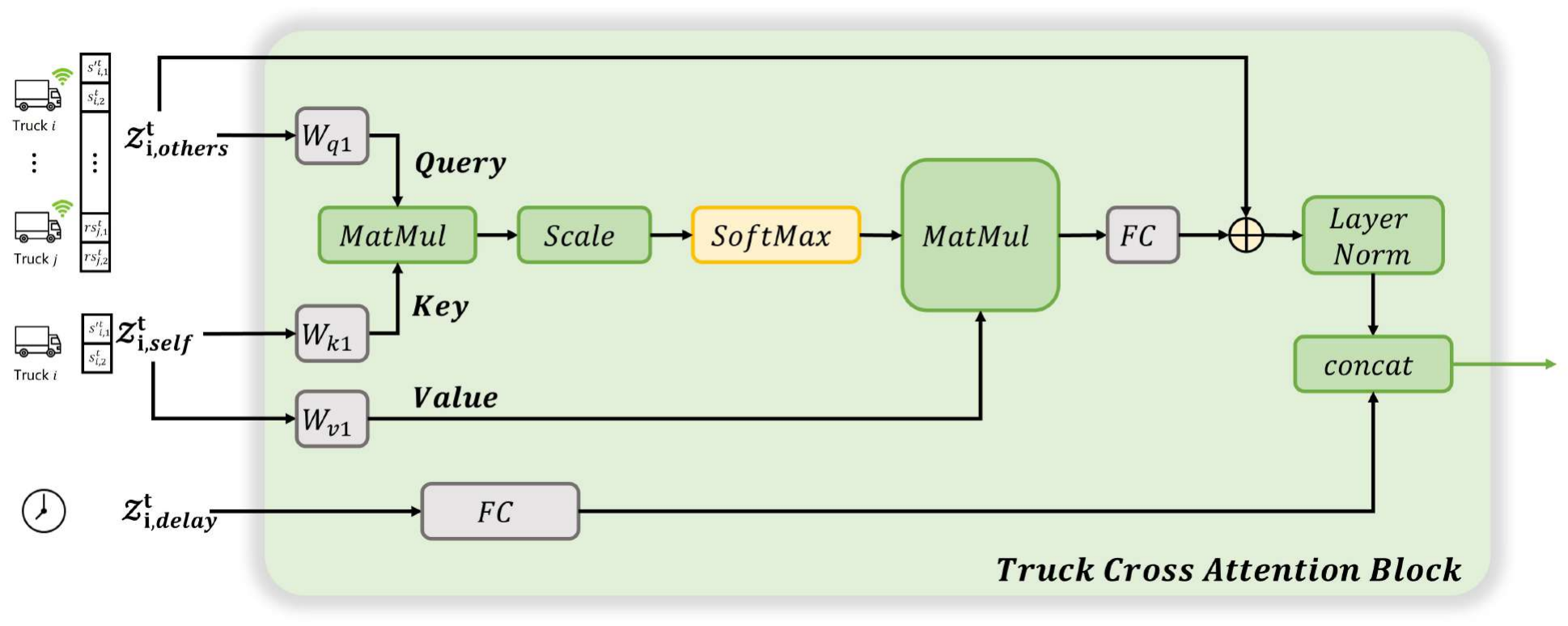}
	\caption{Truck Cross Attention Block.}
	\label{figtca}
\end{figure}

\subsection{Deep Policy Network Architeture}
QMIX \cite{rashid2020monotonic} is a widely recognized deep reinforcement learning algorithm designed to address collaboration tasks in multi-agent scenarios. 
The structure of QMIX cannot well represent the information exchanged between trucks, making it difficult to weigh the fuel consumption and time costs caused by acceleration and deceleration. Hence, directly training the model with QMIX  is hard to converge.
Therefore, we propose TA-QMIX combined with the TCA Block and the TSA Block to solve the training efficiency of platoon coordination policy as shown in Fig. \ref{figtaqmix}.

TA-QMIX comprises $N = |\mathcal{N}|$ individual TCA Agent Networks denoted as $\{\bar{Q}_i\}_{i=1}^N$ with parameters $\{\omega_i\}_{i=1}^N$, and a TSA-Mixing Network denoted as $f$ with parameters $\omega_f$. Similar to the approach in QMIX, TA-QMIX adopts the target network method for training, incorporating target agent networks $\{\bar{Q}'_i\}_{i=1}^N$ with parameters $\{\omega'_i\}_{i=1}^N$, and a target mixing network denoted as $f'$ with parameters $\omega_{f'}$.

\begin{figure}[t]
	\centering
	\includegraphics[scale=0.25]{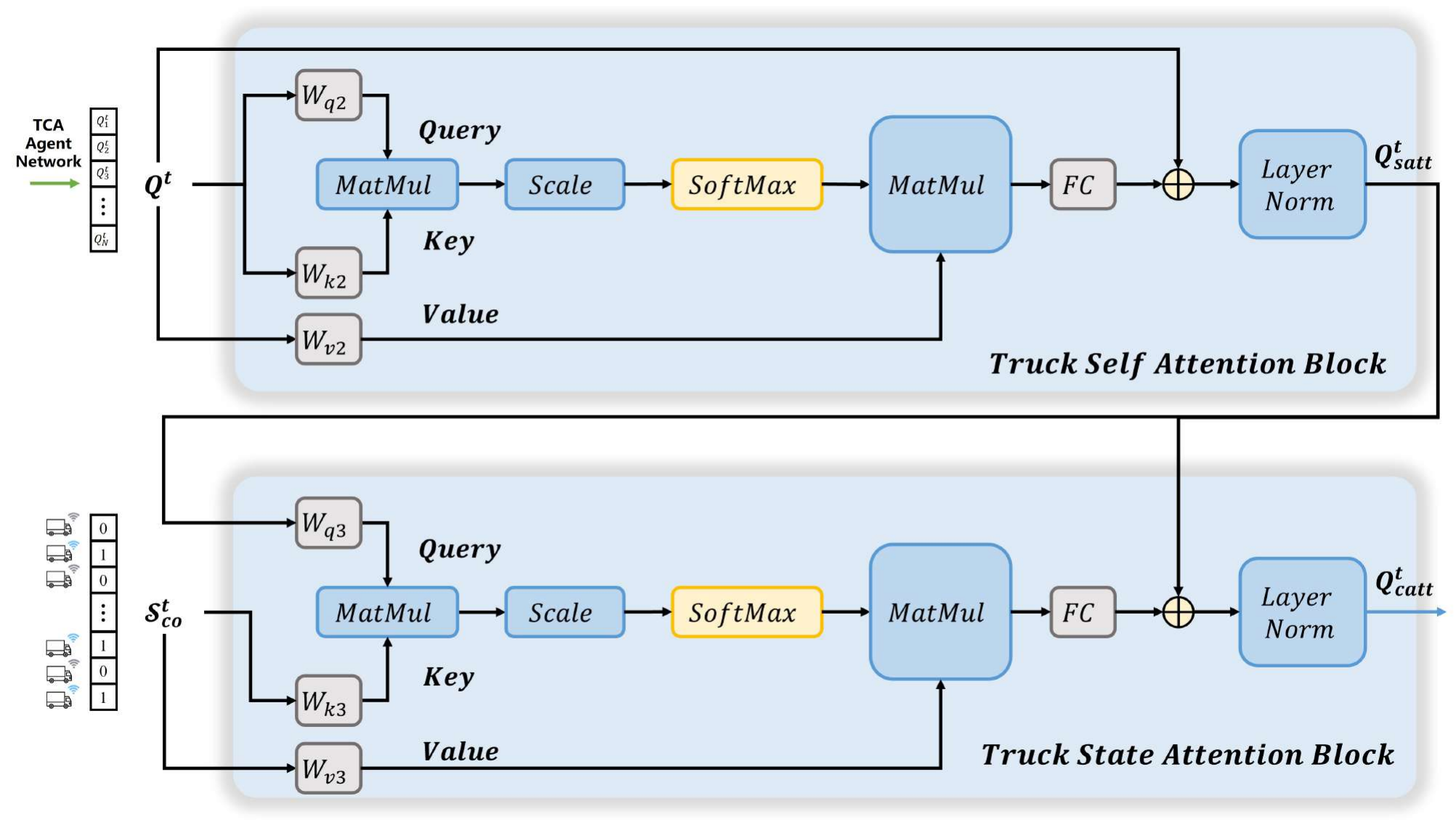}
	\caption{Truck State Attention Block.}
	\label{figtsa}
\end{figure}

\subsubsection{\textbf{Truck Cross Attention Agent Network}} 
TCA Block uses the attention method \cite{vaswani2017attention} to explicitly model the connection between the trucks' information and the observed nearby truck information, calculate the weight of different truck information, and promote cooperation between trucks.
The attention method operates as follows: (i) create three learnable transformation matrices $W_q, W_k, W_v$ embedded within the network; (ii) perform matrix multiplication with the input $S_{in}$ to obtain $Query$, $Key$, and $Value$:
\begin{align}
	Query &= S_{in} \cdot W_q ,\\
	Key &= S_{in}\cdot W_k ,\\
	Value &= S_{in}\cdot W_v. 
\end{align}
(iii) compute the attention coefficient $W_a$ by
\begin{equation}
	W_a = softmax(\frac{Query \cdot {Key}^T}{\sqrt{d_k}}).
\end{equation}
Here, $d_k$ is a constant, which is used to scale the product results. The attention output $S_{att}$ is calculated by:
\begin{equation}
	S_{att} = LN(FC(W_a \cdot Value) + S_{in}),
\end{equation}
where $FC(\cdot)$ represents a linear layer used to the aligned dimensions, $LN(\cdot)$ represents a layer normalization.

Fig. \ref{figtca} shows the architecture of the Truck Cross Attention Block. TCA Block divides the truck information into three parts as input, namely $Query=\mathcal{Z}_{i,other}^t \cdot W_{q1}$, $key=\mathcal{Z}_{i,self}^t \cdot W_{k1}$, $Value=\mathcal{Z}_{i,self}^t \cdot W_{v1}$. 
TCA BLock establishes a weighted connection between its truck and the observation truck by separating its information from the observation information. It allows the truck to pay attention to location information and time information at the same time.

Next, TCA Agent Networks adopt DRQN (Deep Recurrent Q-Network) with Gated Recurrent Unit (GRU) \cite{hausknecht2015deep}, which captures time dependencies between states through recurrent neural networks, enabling trucks to consider previously hidden information $\mathcal{H}_i^{t-1}$ when making decisions and handle continuous decision problems. 
GRU takes $S_{att}^t$, the time encoding vector $FC(\mathcal{Z}_{i, delay}^t)$ and $\mathcal{H}_i^{t-1}$ as input to combine current encoding information and last-time information, i.e.,
\begin{equation}
	\label{gru}
	\mathcal{H}_i^{t} = GRU(FC(S_{att}^t,FC(\mathcal{Z}_{i,delay}^t)), \mathcal{H}_i^{t-1}).
\end{equation}
At each $t$, $\bar{Q}_i$ takes $\mathcal{Z}^t_i$ and $\mathcal{H}_i^{t-1}$ as input to calculate $Q_i^t$ and selects the largest value actions $a_i^t \in A_i(s_i^t)$ that can be selected.
\begin{equation}\label{eqq}
	Q_i^t=\mathop{max}\limits_{a_i^t} \bar{Q}_i(\mathcal{Z}_i^t,\mathcal{H}_i^{t-1}, a_i^t;\omega_i),
\end{equation}
Likewise, the calculation formula of ${Q}'^{t+1}_i$ is denoted as
\begin{equation}\label{eqq'}
	Q'^{t+1}_i=\mathop{max}\limits_{a_i} \bar{Q}'_i(\mathcal{Z}^{t+1}_i,\mathcal{H}'^{t}_i, a_i^{t+1};\omega'_i).
\end{equation}
Especially, $\mathcal{H}_i^{0}$ is an initial zero matrix.  $\mathcal{H}'^{1}_i$ is another independent initial zero matrix for the target network. $Q_i^t$ is the prediction of future rewards for truck $i$ after acting.

\subsubsection{\textbf{Truck State Attention Mixing Network}} Attention mixing network $f$ takes $\mathcal{S}^t=\{\mathcal{S}_{edge}^t,\mathcal{S}_{co}^t\}$ and $\mathbf{Q}^t=\{Q^t_1,Q^t_2,\ldots,Q^t_{N}\}$ as input and $Q_{total}^t$ as output, which can be denoted as
\begin{equation}\label{eqqtotal}
	Q_{total}^t=f(\mathcal{S}^t,\mathbf{Q}^t)=f_2(f_{tsa}(\mathbf{Q}^t,\mathcal{S}_{co}^t),f_1(\mathcal{S}^t_{edge})).
\end{equation}
Here, $Q_{total}^t$ aims to measure the quality of the joint action space of multiple trucks at time $t$. In addition, $f'$ has the same structure as $f$ in (\ref{eqqtotal'}).
\begin{equation}\label{eqqtotal'}
	Q'^{t+1}_{total}=f'(\mathcal{S}^{t+1},\mathbf{Q}^{t+1})=f'_2(f'_{tsa}(\mathbf{Q}^{t+1},\mathcal{S}_{co}^{t+1}),f'_1(\mathcal{S}^{t+1}_{edge})).
\end{equation}

Although $S^t \rightarrow \mathcal{S}_{edge}^t$ makes the size of the state fixed, the information between trucks is lost. It is hard to find with which trucks they may form a platoon based on $\mathcal{S}_{edge}^t$. Hence, 
we add $\mathcal{S}_{co}$ and design the TSA Block, as shown in Fig. \ref{figtsa}, consisting of Truck Self Attention Block and Truck State Attention Block.

The Truck Self Attention Block uses the self-attention method to focus the network on the relationship between the truck's Q-value and others. Truck State Attention Block takes $\mathcal{S}_{co}^t$ as the $Key$ and the $Value$ to explicitly convey cooperative information of trucks for enhancing the estimation of total Q value. Therefore, the output of TSA-Block can be defined as 
\begin{equation}
	\textbf{Q}_{catt}^t = f_{tsa}(\textbf{Q}_i^t, \mathcal{S}_{co}^t).
\end{equation}

As shown in (\ref{eqf11}), $f_1$ aims to encode the state $\mathcal{S}_{edge}^t$ at time $t$ as weights.

\begin{align}
	\label{eqf11}
	f_1(\mathcal{S}^t_{edge}) &= (W_1,B_1,W_2,B_2) \\
	&= (f_{11}(\mathcal{S}^t_{edge}),f_{12}(\mathcal{S}^t_{edge}),f_{13}(\mathcal{S}^t_{edge}),f_{14}(\mathcal{S}^t_{edge})). \notag
\end{align}

Here, $f_{11}(\cdot)$,$f_{12}(\cdot)$, and $f_{13}(\cdot)$ represent the three independent fully connected layers, while $f_{14}(\cdot)$ consists of two fully connected layers and a ReLU activation function.


Equation (\ref{eqf2}) illustrates that $f_2$ multiplies the weights by $\mathbf{Q}^t_{catt}$ calculated by the agent networks.

\begin{equation}
	\label{eqf2}
	\begin{split}
		f_2(\mathbf{Q}^t_{catt},W_1,B_1,W_2,B_2)= \\
		|W_2|\cdot \text{ELU}(|W_1|\cdot \mathbf{Q}^t_{catt}+B_1) +B_2,
	\end{split}
\end{equation}
where $\text{ELU}(\cdot)$ is the ELU activation function. $|\cdot|$ is the absolute activation function, which is to ensure $W_1$ and $W_2$ are non-negative. By adding absolute value constraints to the weights, the contribution of each truck's action to the global joint action is qualified(i.e., each truck tends to cooperate with the other).

\subsection{Training Algorithm}
The entire training process can be divided into experience data generation and network training, which alternate in execution. We summarize the pseudo-code of TA-QMIX in Algorithm 1.

\renewcommand{\thealgorithm}{1}
\begin{algorithm}
	\caption{Training Process of TA-QMIX Algorithm}
	\begin{algorithmic}[1]\label{alg2}
		\Require $N$ trucks in the transport network, Replay buffer $\mathcal{D}$, episode length $T_e$, joint action space $\mathbf{a}$, batch size $B$, discount factor $\gamma$, target network update frequency $C$, greedy policy $\epsilon$.
		\Ensure Agent-networks $\{\bar{Q}_i\}_{i=1}^N$ and attention mixing network $f$ characterized by $\{w_i\}_{i=1}^N$ and $w_{f}$ respectively.
		\State \textbf{Initialize} Agent-networks $\{\bar{Q}_i\}_{i=1}^N$ with random weights $\{w_i\}_{i=1}^N$, target Agent-networks $\{\bar{Q}'_{i}\}_{i=1}^N$ with weights $\{w'_{i}\}_{i=1}^N \gets \{w_i\}_{i=1}^N$, attention mixing network $f$ with random weights $w_{f}$, target attention mixing network $f'$ with weights $w_{f'} \gets w_{f}$.
		\For{each training step}
		\For{$t= 1$ to $T_e$}
		\State Obtain observation $\mathcal{Z}^t$ and state $\mathcal{S}^t$.
		\State Sample joint actions ${A}_t$ from the TCA Agent Networks with $\epsilon$-greedy policy.
		\State Execute ${A}^t$ and observe rewards $R^t$, next joint observation $\mathcal{Z}^{t+1}$ and next state $\mathcal{S}^{t+1}$.
		\EndFor
		\State Store transition $(\mathcal{Z}^{1:T_e+1}, {A}^{0:T_e}, \mathcal{S}^{1:T_e+1}, R^{1:T_e})$ in $\mathcal{D}$.
		\If{$|\mathcal{D}| \geq B$}
		\State Choose min-batch of transitions with B samples in $\mathcal{D}$.
		\For{$t = 1$ to $T_e$}
		\For{$b = 1$ to $B$}
		\State Calculate each truck's local Q-values $\mathbf{Q}^t_b$  and local target Q-values $\mathbf{Q}'^{t+1}_b$ by (\ref{eqq}) and  (\ref{eqq'}).
		\State Calculate total Q-value $Q^{t}_{total_{b}}$ and target Q-value $Q'^{t+1}_{total_b}$ with $\mathbf{Q}^t_b$ and $\mathbf{Q}'^{t+1}_b$ by (\ref{eqqtotal}) and (\ref{eqqtotal'}).
		\EndFor
		\EndFor
		\State Calculate TA-QMIX loss by (\ref{eqloss}).
		\State Perform a gradient descent step on $\mathcal{L}(\{w_i\}_{i=1}^N,w_f)$ with respect to the parameters $\{w_i\}_{i=1}^N$ and $w_{f}$.
		\EndIf
		\State Every $C$ train steps reset $\{w'_{i}\}_{i=1}^N \gets \{w_i\}_{i=1}^N$, $w_{f'} \gets w_{f}$.
		\EndFor
	\end{algorithmic}
\end{algorithm}

\subsubsection{\textbf{Experience Data Generation}}
During experience data generation, each truck $i \in \mathcal{N}$ continuously makes decisions based on its current $\bar{Q}_i$ until all trucks reach the endpoint (complete an episode). The generated data is then stored in the experience replay buffer $D$. The process of experience data generation can be found in (Algorithm 1, lines 3-8).

\subsubsection{\textbf{Network Training}}
After data generation, $B$ episode data is randomly selected from the buffer for training.
According to (\ref{eqq}) and (\ref{eqq'}), for $t \in \{1,2,\ldots,T_e\}$ and $b \in \{1,2,\ldots,B\}$, the agent networks and the target networks calculate $\mathbf{Q}^t_b$ and $\mathbf{Q}'^{t+1}_b$ from the data. Then $f$ and $f'$ output the total Q-value ${Q}_{total_b}^t$ and ${Q}'^{t+1}_{total_b}$ according to (\ref{eqqtotal}) and (\ref{eqqtotal'}). The process can be found in (Algorithm 1, lines 9-14).



Finally, TA-QMIX is trained by the following loss:
\begin{equation}
	\mathcal{L}(\{\omega_i \}_{i=1}^N,\omega_f)=\sum_{b=1}^{B}{(R^{1:T_e}+\gamma {Q}'^{2:T_e+1}_{total_b}-{Q}^{1:T_e}_{total_b})^2}. \label{eqloss}
\end{equation}
Here, $R^{1:T_e}$ represents the reward from time $1$ to $T_e$, and $\gamma$ represents the discount factor.
The training efficiency will be demonstrated in the experimental results.

After the training process, a coordination policy can be implemented with the trained Q-network. 
Each truck can use the trained model to adaptively and autonomously make decisions throughout the traveling journey. Algorithm 2 illustrates the process.

\renewcommand{\thealgorithm}{2}
\begin{algorithm}
	\caption{Execute Process of Single Truck}
	\begin{algorithmic}[1]\label{alg3}
		\State \textbf{Initialize} truck $i$'s agent-network $\bar{Q}_i$ with the trained weights $w_i$, state $s_i^0$, action space $a_i^0$, departure time $d_i$, a fixed set of hubs $\mathcal{V}_i=\{i_1,i_2,\ldots,i_{n_i}\}$, initial hidden matrix $\mathcal{H}_i^{0}$.
		\For{$t= 1$ to $T_e$}
		\If{ $t >= d_i \land s_{i,1}^t \neq i_{n_i}$}
		\State Obtain observation $\mathcal{Z}_i^t$ and its state $s_i^t$.
		\State $a^t_i=\mathop{argmax}\limits_{a_i^t}\bar{Q}_i(\mathcal{Z}^t_i,\mathcal{H}_i^{t-1},a_i^t;\omega_i)$.
		\State Calculate $\mathcal{H}_i^t$ by (\ref{gru}).
		\State $s^{t+1}_i \gets T_i(s_i^t,a_i^t)$.
		\EndIf
		\EndFor
	\end{algorithmic}
\end{algorithm}

\section{Experiments}
In this section, we conduct experiments on the transportation network in the Yangtze River Delta region of China to illustrate the efficiency of the proposed method. 
\begin{figure}[h]
	\centering
	\includegraphics[scale=0.15]{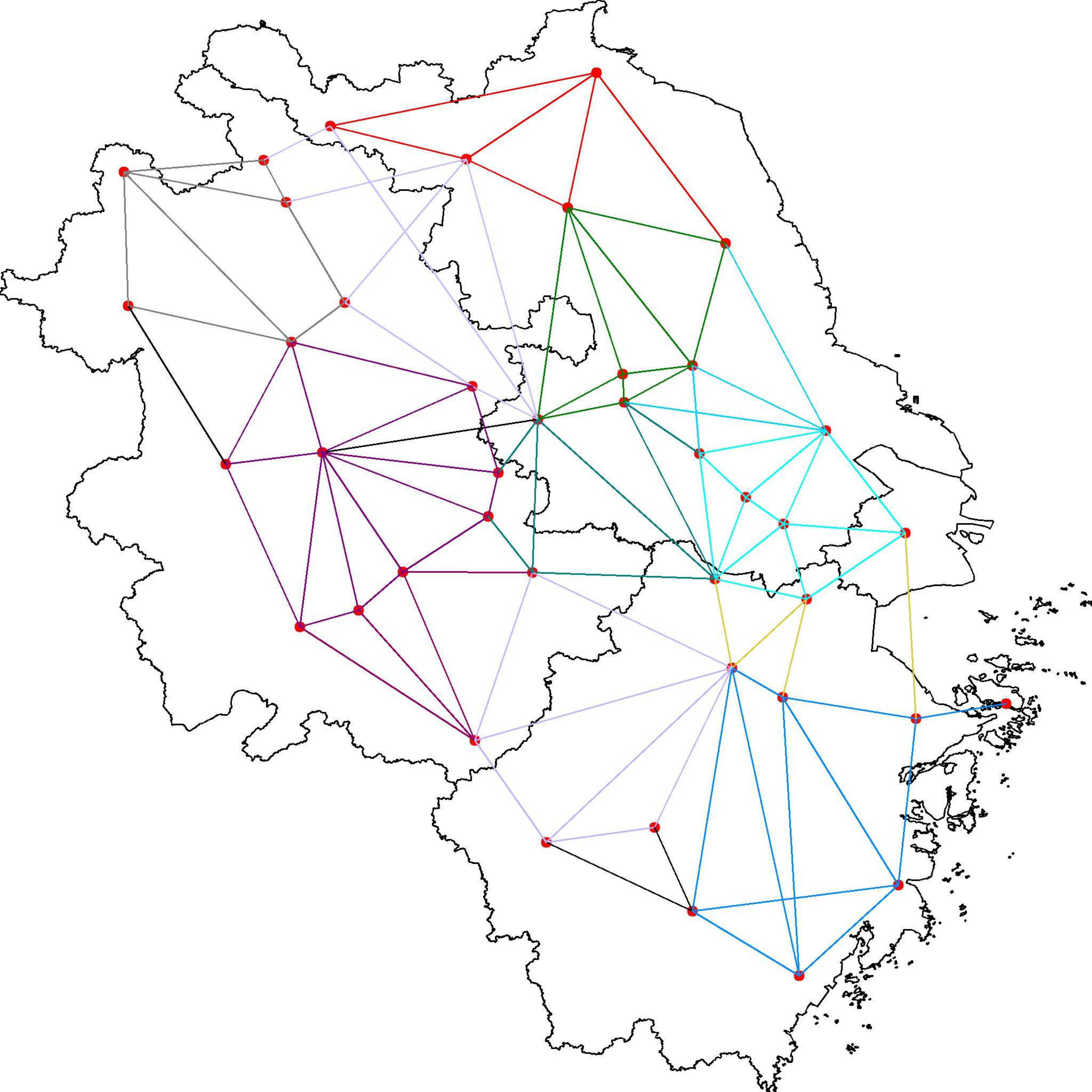}
	\caption{Transportation Network Map of Yangtze River Delta Region. We divided the transportation network into 12 blocks with different colors to generate freight missions.}
	\label{fig4}
\end{figure}

\subsection{Experiment Setting}
Our experiments are implemented with an Intel(R) Xeon(R) Gold 6240R CPU @ 2.40GHz, 4 NVIDIA 3090 GPUs, and Pytorch. Illustrated in Fig. \ref{fig4}, we built a transportation network simulation environment in the Yangtze River Delta to simulate large-scale truck freight scenarios. The simulation environment allows for arbitrarily adjusting the number of trucks, departure time, arrival time, driving path, and freight targets (lower fuel or on-time arrival).
The transportation network spans across Shanghai, Zhejiang, Jiangsu, and Anhui provinces, encompassing 41 hubs, each corresponding to a city, and 202 directed edges with an average length of $140.95km$. The transportation network data is sourced from Amap\footnote{City locations and road information are obtained from www.amap.com/}.
A demonstration video of the experiment is available\footnote{\url{https://drive.google.com/file/d/1rNgGhgHWla9baFYOcq1q6NQ1Tk5lMhuW/view?usp=sharing}}.
\subsubsection{Simulation Setup}
We set $|\mathcal{N}|=100$ and assume that each truck's starting and ending points follow the shortest fixed driving route, with departure times randomly selected between 5:00 am and 11:00 am. Let $t = 1$ start from 5:00 am, time step $\Delta t = 4$ minutes, and maximal time $T_e=1000$ minutes.
During the entire journey, we assume that the action $v_l=60 km/h$,  $v_m==75 km/h$, $v_h=90 km/h$ and the noise $\xi \sim 0.01 * N(0,1)$. 
The maximum allowable delay time rate is $\mathcal{K}_d = 1.1$. 
Similar to \cite{liang2015heavy}, the air drag coefficient is set as $\phi_s=1$ and $\phi_p=0.68$.
Each edge of the transportation network is divided into $\alpha =10$ segments and observation parameters are set as $q=5$.
\subsubsection{Freight Mission Generation}
Considering the efficiency of the training samples, as shown in  Fig. \ref{fig4}, we divided the transportation network into 12 blocks with different colors, each with approximately 5-9 hubs, and limited the truck freight mission to each block. The reduced transportation topology map can greatly increase the traffic flow in the simulation environment, giving trucks more opportunities to adjust their speed to form a platoon, which is more conducive to network training and convergence.

Similar to the method used in \cite{johansson2021strategic}, we randomly generated 5 freight missions for each block, totaling 60 freight missions for training. The information for each freight mission includes the starting point, ending point, driving route, and departure time of 100 trucks. The probability of generating the starting and ending points is directly proportional to the population of the city.

Due to different freight missions and topological maps leading to different total times $t_{total}$, according to \eqref{rewardt}, \eqref{rewarde} and \eqref{rewardf}, this would result in trucks in the same states making the same action under different scenarios receiving different rewards, which is not conducive to the network's prediction of Q-values. Therefore, we replace $t_{total}$ in the formula with the average value of $t_{avgtotal}$ across the aforementioned 60 freight scenarios. At the same time, using $w_1$ and $w_2$, we scale the reward proportionally to the $t_{avgtotal}$.
In this experiment, the average value of $t_{avgtotal}=18028.7167 $ minutes and we set $w_1 = 7.03125 * t_{avgtotal}$ and $w_2 = 10.5 * t_{avgtotal}$.

\subsubsection{Training Parameter}
In Algorithm 2, we set the number of training steps as $8000$, the batch size $B=16$, the experience replay buffer size $|\mathcal{D}|=100$, $\gamma=0.99$, learning rate $5e-4$, and let the target network update every $C=50$ epoch. We set the initial value of the greedy policy $\epsilon$ to be $1$ and the minimum value as 0.05, and let it decrease by 0.00019 each time step.

\subsection{Baselines and Metrics}
The proposed method is compared with the following baseline method, including the no-coordination and Nash equilibrium-based coordination method.
\begin{itemize}
    \item \textbf{No Coordination (NC):} NC refers to trucks that are not guided by any coordinated strategy. The truck travels at a medium speed $v_m$ throughout the journey, without any waiting at the hub, relying solely on the same departure time to form a platoon.
    \item \textbf{Nash equilibrium (NE)-based coordination:} NE-based method \cite{johansson2021strategic} considers each truck as a non-cooperative agent and uses potential games to calculate the Nash equilibrium solution for all trucks in the large-scale transportation network. This method essentially uses the potential game to maximize each truck's benefit at the decision-making moment of the transportation network. 
    Using settings similar to \cite{johansson2021strategic}, trucks travel at a fixed speed $v_m$ and can wait for other trucks to form a platoon at the hub. We implement the NE-based method in a no-noise simulator (i.e. we don't consider the state transition noise $\xi$ in the no-noise simulator) with a calculation interval of $5$ minutes and a node number of $H=1$.
    \item \textbf{QMIX:} QMIX uses the same training parameters as our proposed method for training.
\end{itemize}

The evaluation indicators need to be able to accurately evaluate the impact of a series of truck behaviors on fuel, time, and cooperation rate during the journey.
In this paper, we use the following indicators to evaluate coordination performance.
\begin{itemize}
    \item \textbf{Fuel Saving Rate $F_r$:} According to \eqref{fuelrate}, the fuel-saving rate is calculated as the ratio of the total fuel consumed to the fuel consumed by the NC method, measuring the impact of the platoon coordination strategy on the overall system fuel savings.
        \begin{equation}
            F_r = \sum_{t=1}^{T_e}\sum_{i\in \mathcal{N}} J_{i,a}(s_i^t,a^t_i)
        \end{equation}
    \item \textbf{Average Delay Time $T_d$} and \textbf{Delay Time Rate $T_r$:} The delay time rate calculates the proportion of the sum of truck delay times to the total travel time, and measures the impact of platoon coordination strategies on truck delayed arrival. $T_d$ and $T_r$ can be defined as
        \begin{equation}
        	T_d = \frac{1}{|\mathcal{N}|}\sum_{i \in \mathcal{N}}{t_i^a(i_{n_i})-t'^a_i(i_{n_i})},
        \end{equation}
        
        \begin{equation}
        	T_r = \frac{\sum_{i \in \mathcal{N}}{t_i^a(i_{n_i})-t'^a_i(i_{n_i})}}{t_{total}}.
        \end{equation}
    
    \item \textbf{Platoon Journey Rate $P_j$:} Platoon journey rate calculates the proportion of the distance traveled by trucks in the platoon to the total distance traveled by the system. It can directly reflect the efficiency of truck cooperation and indirectly reflect fuel savings. Platoon journey rate $P_r$ can be calculated by
        \begin{equation}
            P_j = \frac{\sum_{t=1}^{T_e}\sum_{i\in \mathcal{N}}{\delta_i(S^t) a_i^t \Delta t + \xi_i^t}}{\sum_{i\in \mathcal{N}}\sum_{k=1}^{n_{i-1}}{L(i_k,i_{k+1})}}.
        \end{equation}
        Here, $\delta_i(S^t)$ can be defined as (\ref{deltas}) representing whether the truck $i$ is in a platoon.
        \begin{equation}
            \label{deltas}
            \delta_i(S^t) = 
            \begin{cases}
                1, & \text{if } \Omega_i(S^t) \neq \varnothing, \\
                0, & \text{otherwise}.
            \end{cases}
        \end{equation}
    
    \item \textbf{Function Value $F_v$:} Function value calculates the function value of problem \eqref{IO_formulation}. Given weights $w_1$ and $w_2$, the objective function value that the model can obtain measures the balance utility of the model on fuel consumption and time delay.
        \begin{equation}
            F_v = \sum_{i\in \mathcal{N}} \sum_{t=1}^{T_e} w_{1} {J_{i,f}(S^t,v^t)} +  w_{2} J_{i,d}(S,v).
        \end{equation}
    \item \textbf{Timeout Rate $T_o$:} Timeout rate calculates the proportion of trucks in the platoon coordination strategy whose travel time exceeds the given time budget ratio $\mathcal{K}_d$. It and the time delay rate can effectively reflect the strategy's control over the time budget.
        \begin{equation}
            T_o = \frac{1}{|\mathcal{N}|}|\{j:j\in \mathcal{N}, \frac{t_j^a(j_{n_j}) - d_j}{t_j^{a^\prime}(j_{n_j}) - d_j} > \mathcal{K}_d\}|.
        \end{equation}
    \item \textbf{Participation Rate $P_r$:} Participation rate quantifies the proportion of trucks that form a platoon during the journey. It can reflect the participation of trucks in a coordination scenario of the current strategy and measure the cooperation rate of trucks.
        \begin{equation}
            P_r = \frac{1}{|\mathcal{N}|}|\{j:j\in \mathcal{N}, \exists t \in [1,T_e], \delta_j(S^t) = 1\}|.
        \end{equation}
\end{itemize}

Specifically, since the NE-based method considers the tolerated delay time as a constraint for calculation, it will not allow the truck to arrive over time, so we do not evaluate the timeout rate $T_o$ of the NE-based method.

\subsection{Experimental Result and Analysis}
\begin{table*}[t]
\label{comparetable}
\caption{Comparative Experimental Results in 60 Freight Missions.}
\centering
\resizebox{0.99\textwidth}{!}{
    \begin{tabular}{cccccccc}
    
    \toprule
    Coordination Method & $F_r[\%] \uparrow $     & $T_d[min]\downarrow$  &$T_r[\%]\downarrow$   & $P_j[\%] \uparrow$     & $F_v \uparrow$     & $T_o[\%]\downarrow$     & $P_r[\%] \uparrow$     \\
    \cmidrule(r){1-8}
    NC                  & $1.30 \pm 0.56$ & $-$ & $-$ & $5.26 \pm 3.32$ & $1113.08 \pm 702.47$ & $-$ & $6.32 \pm 3.66$\\
    NE-based               & $4.26 \pm 1.35$ & \boldmath{$1.00 \pm 0.48$} & \boldmath{$0.54 \pm 0.21$} & $25.33 \pm 7.41$ & $4370.23 \pm 1385.74$ & $-$ & $31.70 \pm 10.30$\\
    QMIX                 & $11.00 \pm 0.89$ & $8.64 \pm 1.79$ & $4.80 \pm 0.69$ & $24.44 \pm 6.17$ & $3966.23 \pm 1491.29$ & $11.42 \pm 4.00$ & $35.73 \pm 8.14$\\
    TA-QMIX             & \boldmath{$13.56 \pm 1.14$} & $10.16 \pm 2.42$ & $5.58 \pm 0.61$ & \boldmath{$30.53 \pm 6.76$} & \boldmath{$5962.79 \pm 1765.53$} & \boldmath{$0.22 \pm 0.45$} & \boldmath{$49.39 \pm 10.56$}\\ \bottomrule
    \end{tabular}}
\end{table*}

\begin{figure}[t]
	\centering
	\includegraphics[scale=0.45]{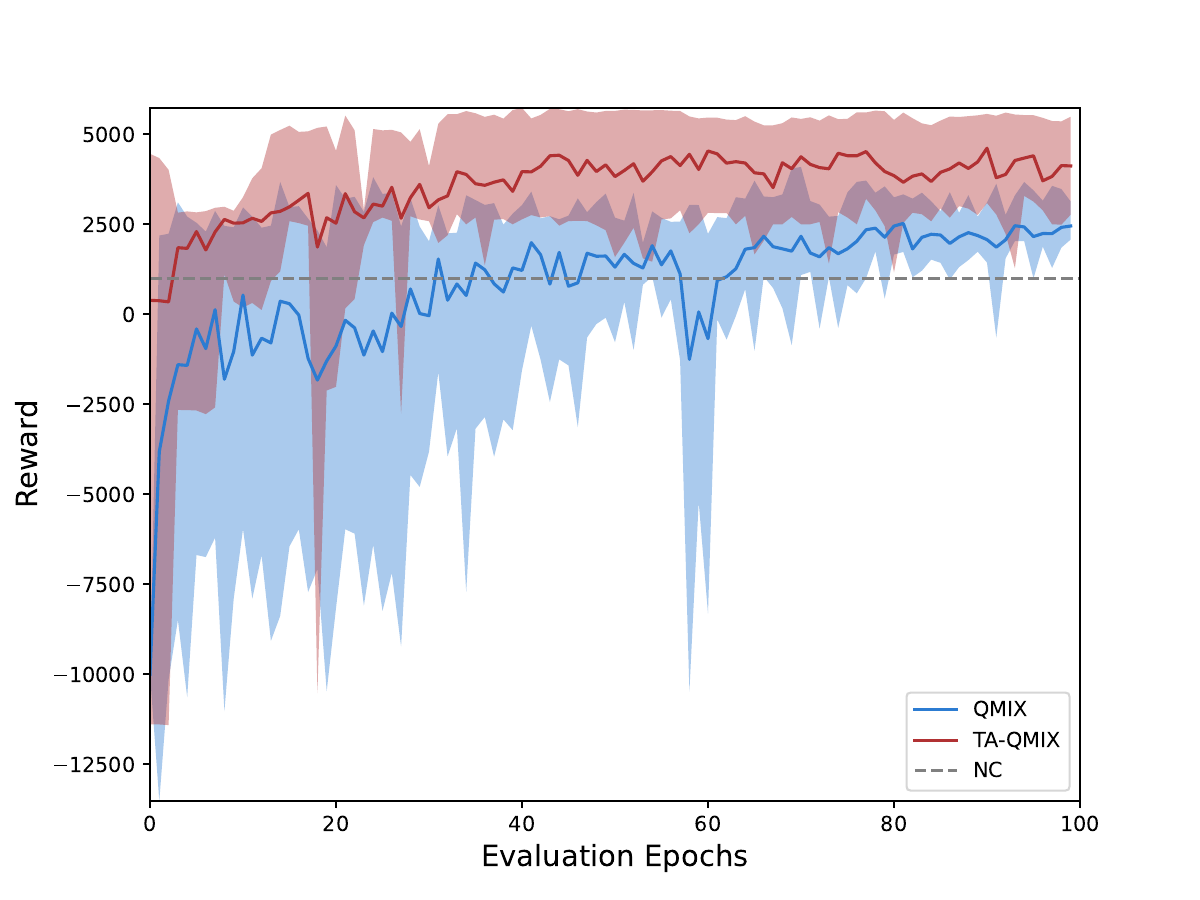}
	\caption{Training process.}
	\label{figreward}
\end{figure}

\subsubsection{Comparison Experiment}
We use the above 60 freight missions as the training set to train our platoon coordination policy. We train this policy for $4000$ epochs and test it every $40$ epochs. 
Fig. \ref{figreward} shows the training process of QMIX and TA-QMIX with an average taken from 5 repeated training.
The gray lines represent the average rewards obtained using the NC method in 60 freight missions. 
Due to the random state transition noise $\xi$, we repeat the test $3$ times for each mission. 
From Fig. \ref{figreward}, it can be observed that QMIX exhibits a high variance in exploration during the initial training phase and eventually stabilizes at a reward value slightly higher than that achieved without any strategy. In contrast, TA-QMIX quickly discovers better strategies early in the training process and continues to improve, ultimately stabilizing at a significantly higher reward value. This indicates that TA-QMIX is more effective in rapidly identifying superior truck coordination strategies. 
Specifically, since the NE-based method directly calculates $F_v$ for each freight mission rather than using the reward to coordinate trucks, it is not compared in the figure.

After training to obtain the best platoon coordination strategy, we compare it with the baseline methods as shown in Table II. 
Compared with three baselines, TA-QMIX achieves a higher fuel saving rate $F_r$, platoon journey rate $P_j$, participation rate $P_r$, and function value $F_v$. 
The NE-based method achieves lower time delay performance. However, due to the lack of speed regulation, it misses many opportunities for platoon formation, which prevented it from saving more fuel and achieving higher rewards.
Although TA-QMIX sacrifices some time cost to adjust the speed to form a platoon, most trucks reached the ending point within a tolerable time. Among the 60 freight missions, on average only $2$ trucks may exceed the time limit.
This indicates that the proposed method can enable trucks to plan their driving speed appropriately during the journey, form more platoons, and save fuel as much as possible without exceeding the time limit.

\begin{figure}[t]
	\centering
	\includegraphics[scale=0.45]{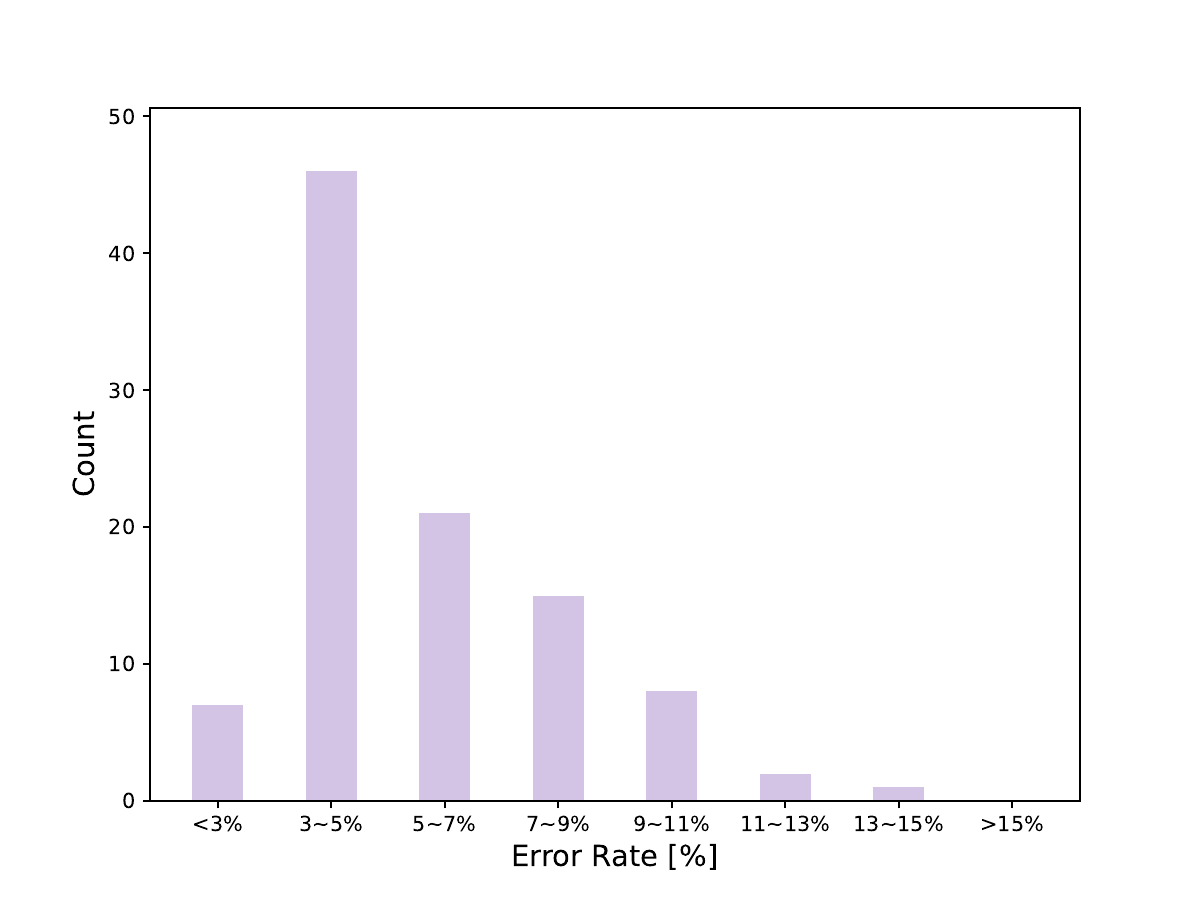}
	\caption{The Error rate of reward and function value.}
	\label{figerror}
\end{figure}

\subsubsection{Reward and Function Value}
According to \eqref{eqtimedecompose}, we approximately decompose the original platoon coordination problem \eqref{IO_formulation}. And in the experiment, we use $t_{avgtotal}$ to substitute for $t_{total}$. Those will lead to a small error between the original function value $F_v$ and the reward.
Thus, in Fig. \ref{figerror}, we use \eqref{eqerror} to demonstrate the error rate $E_r$ between each test reward and function value $F_v$ in the training process. 
\begin{equation}
    \label{eqerror}
    E_r = \frac{|R - F_v|}{F_v}
\end{equation}
Fig. \ref{figerror} illustrates the error rate between the 100 test rewards and function values during the TA-QMIX training process. Most error rates fall within a low range of 3\% to 5\%, indicating that the strategies obtained by MADRL approximate the original problem effectively.


\begin{table*}[t]
\label{comparetable}
\caption{Generalization Experimental Results in 60 Testing Freight Missions.}
\resizebox{0.99\textwidth}{!}{
\centering
    \begin{tabular}{cccccccc}
    
    \toprule
    Coordination Method & $F_r[\%] \uparrow $     & $T_d[min]\downarrow$  &$T_r[\%]\downarrow$   & $P_j[\%] \uparrow$     & $F_v \uparrow$     & $T_o[\%]\downarrow$     & $P_r[\%] \uparrow$     \\
    \cmidrule(r){1-8}
    NC                  & $1.41 \pm 0.56$ & $-$ & $-$ & $5.93 \pm 3.43$ & $1242.12 \pm 719.02$ & $-$ & $7.73 \pm 3.63$\\
    NE-based               & $4.50 \pm 1.37$ & \boldmath{$1.07 \pm 0.55$} & \boldmath{$0.56 \pm 0.21$} & $26.84 \pm 7.78$ & $4635.25 \pm 1413.55$ & $-$ & $34.25 \pm 9.53$\\
    TA-QMIX             & \boldmath{$13.71 \pm 1.07$} & $10.28 \pm 2.56$ & $5.58 \pm 0.63$ & \boldmath{$30.89 \pm 6.91$} & \boldmath{$6164.13 \pm 1831.88$} & $0.25 \pm 0.47$ & \boldmath{$50.02 \pm 9.27$}\\ \bottomrule
    \end{tabular}}
\end{table*}

\begin{figure*}[t]
	\centering
        \subfloat[Platoon journey rate $P_j \uparrow$]{
		\includegraphics[scale=0.28]{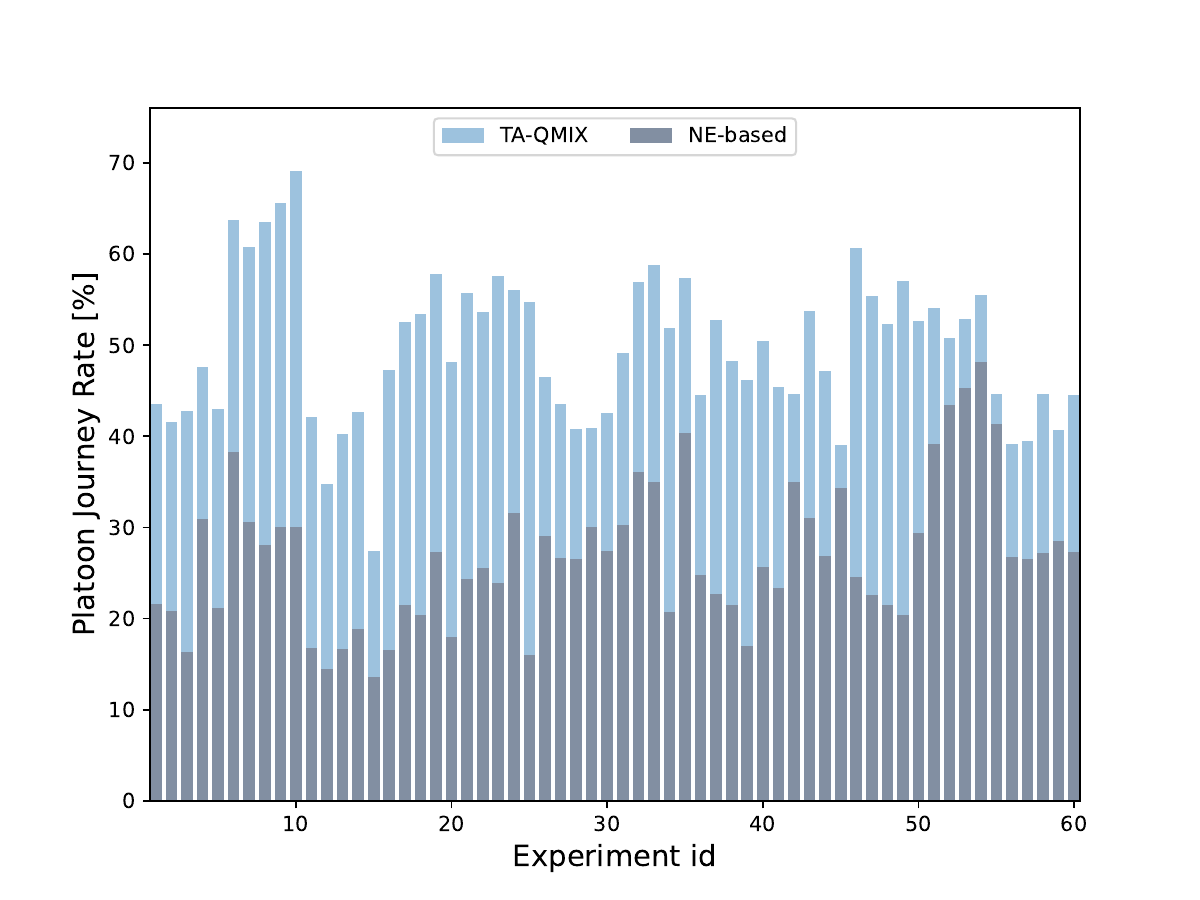}
        \label{figevaluatea}}
        \subfloat[Fuel saving rate $F_r \uparrow$]{
		\includegraphics[scale=0.28]{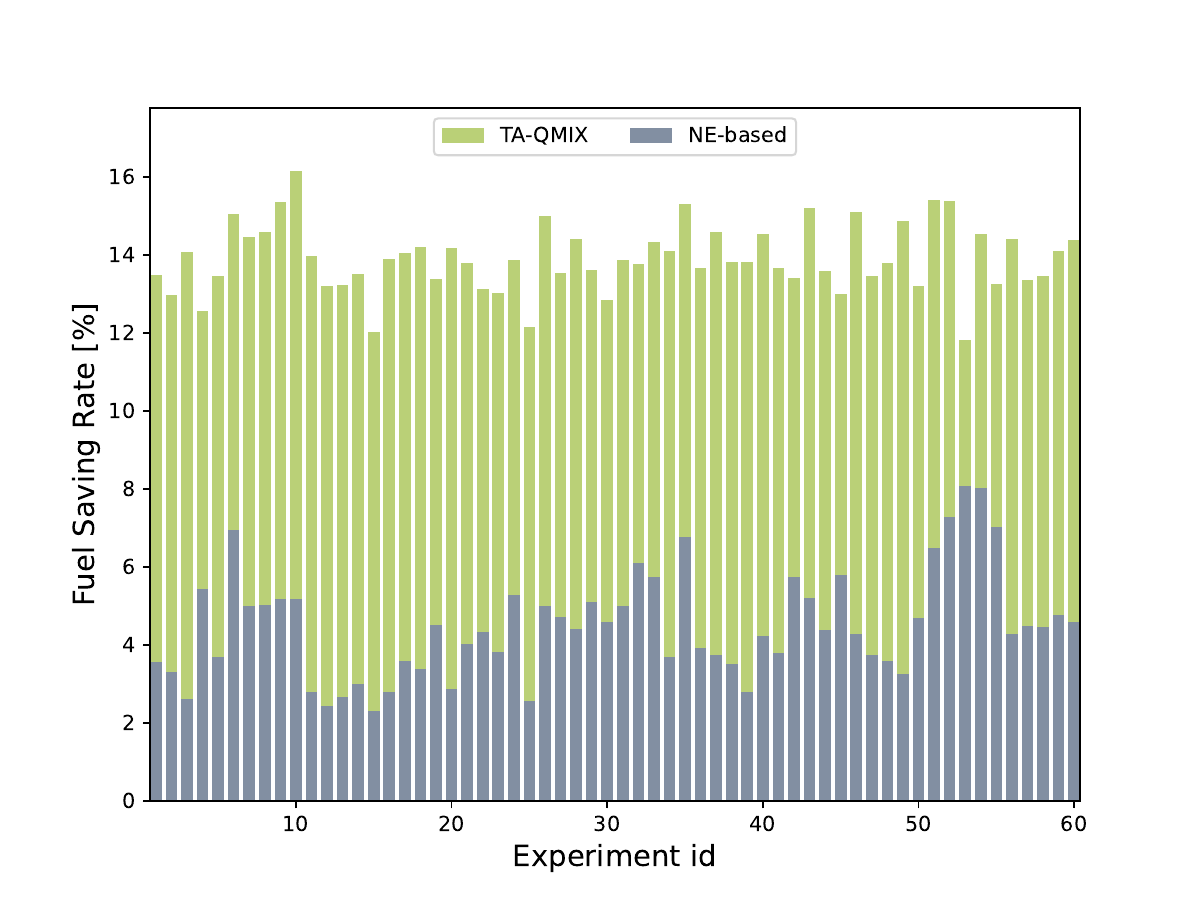}
        \label{figevaluateb}}
        \subfloat[Function value $F_v \uparrow$]{
		\includegraphics[scale=0.28]{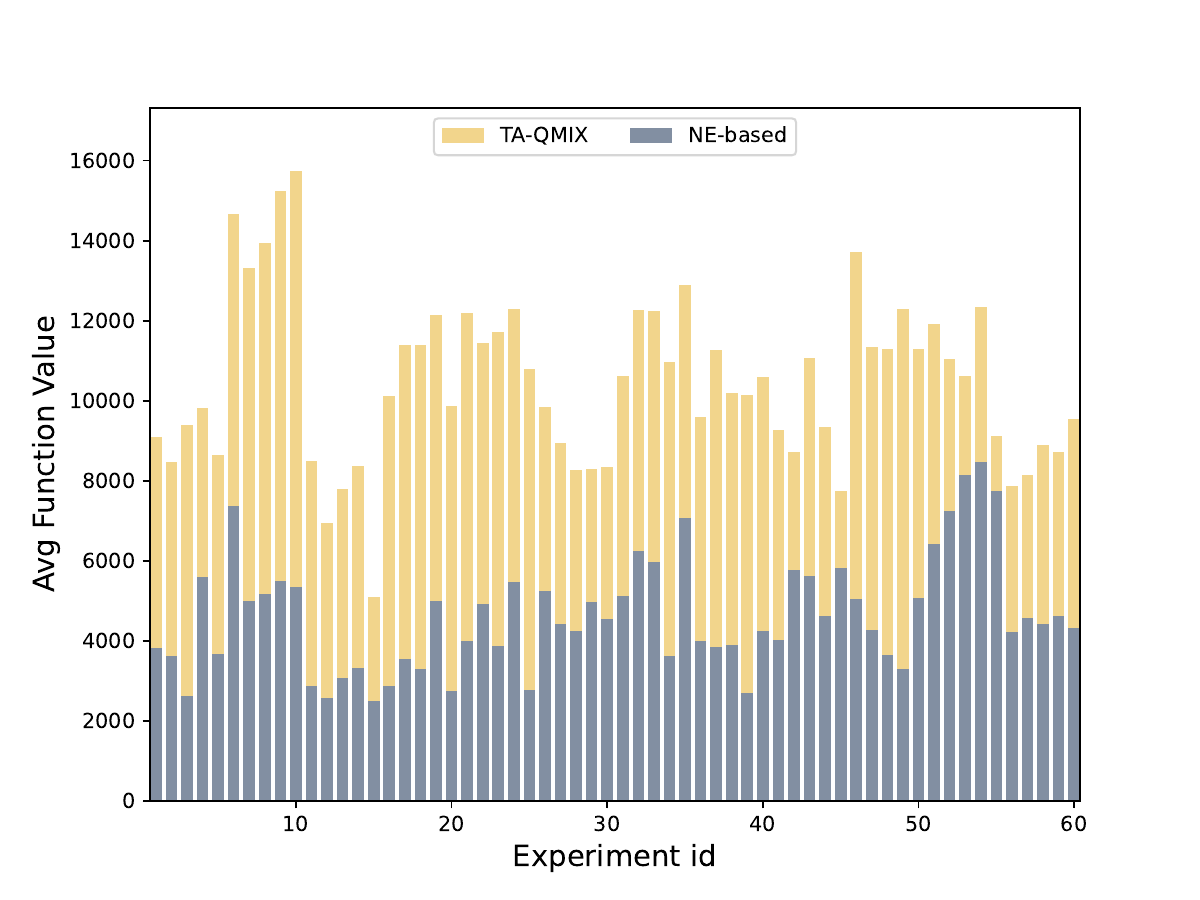}
	\label{figevaluatec}}\\
        \subfloat[Participation rate $P_r \uparrow$]{
		\includegraphics[scale=0.28]{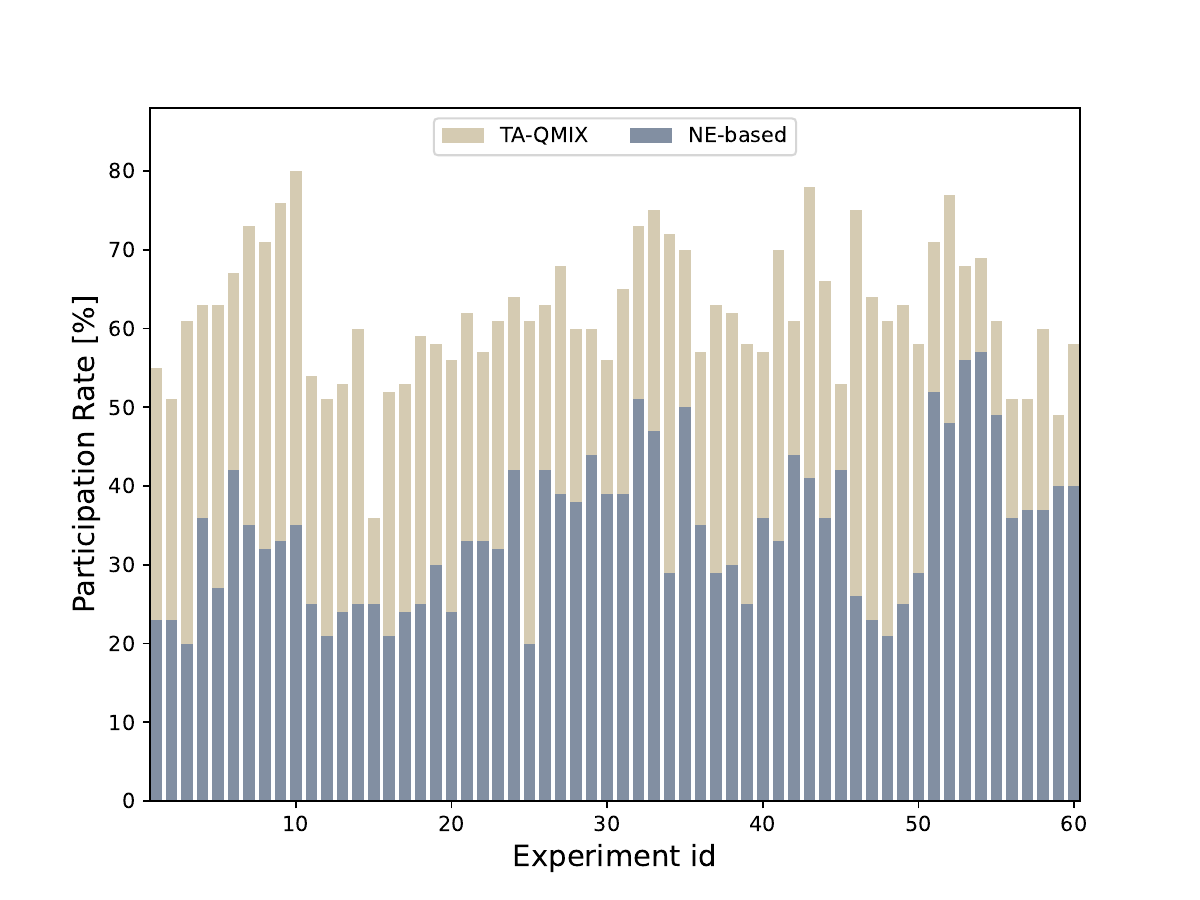}
        \label{figevaluated}}
        \subfloat[Avg delay time $T_d \downarrow$]{
		\includegraphics[scale=0.28]{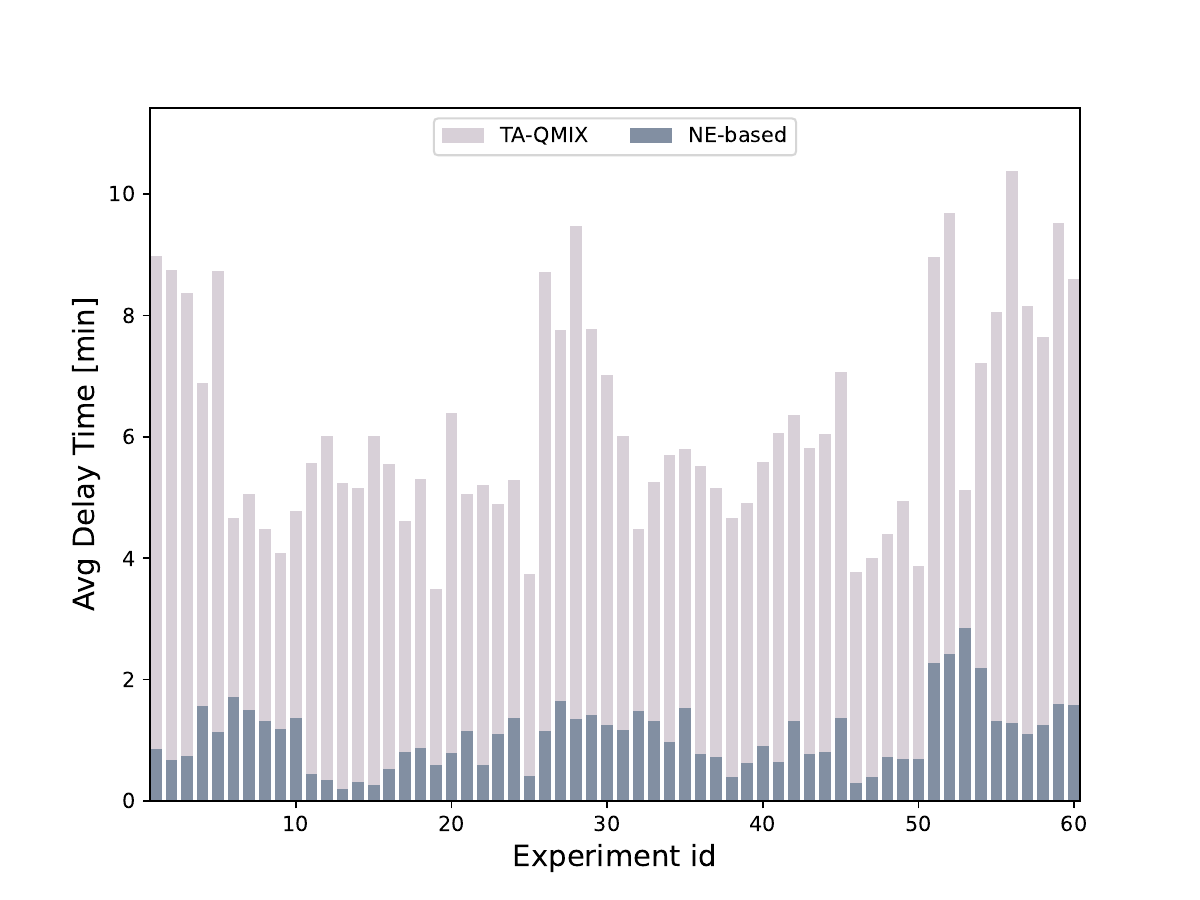}
        \label{figevaluatee}}
        \subfloat[Avg delay time rate $T_r \downarrow$]{
		\includegraphics[scale=0.28]{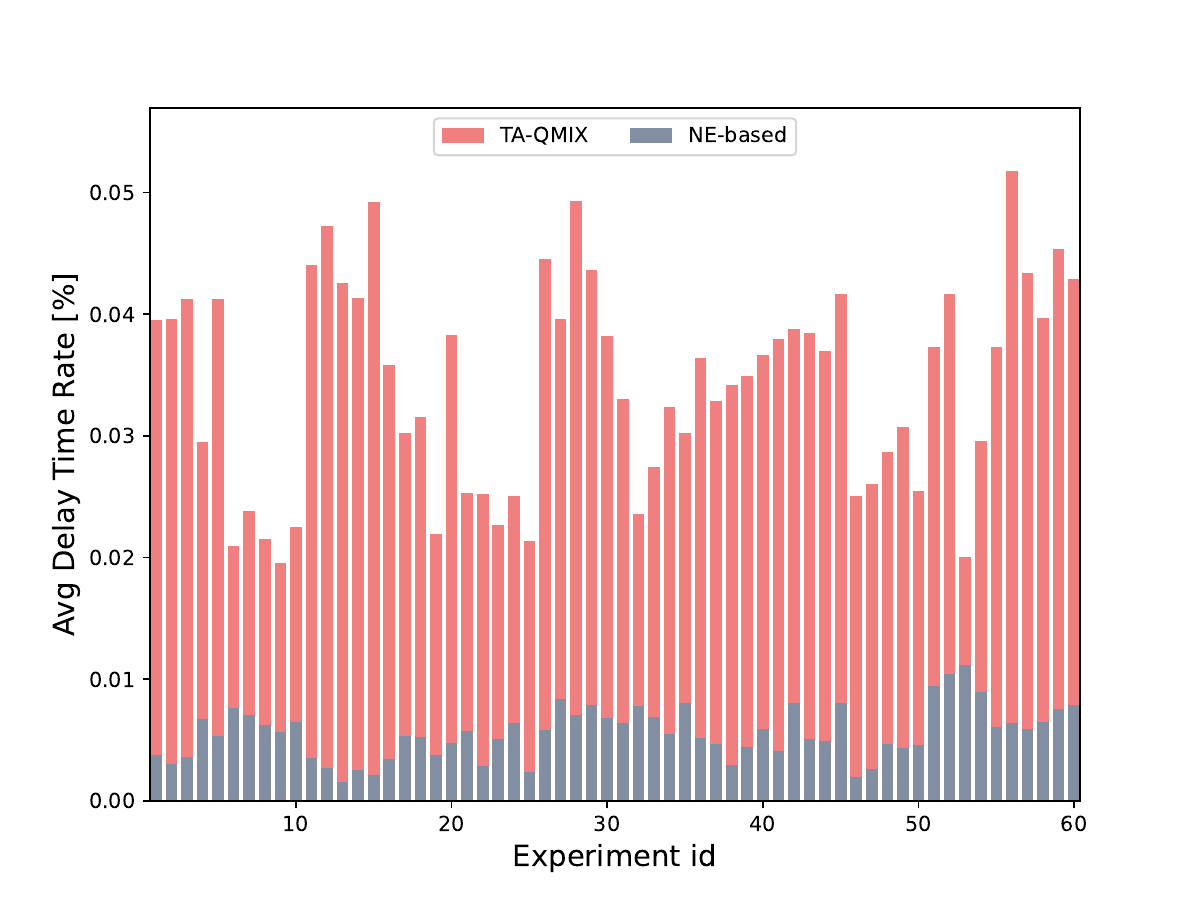}
	\label{figevaluatef}}
        \caption{Testing set comparison results.}
	\label{figevaluate}
\end{figure*}

\subsubsection{Generalization experiment}
For evaluating the generalization performance of our method, we generate an additional 60 freight missions using the same method as the testing set. The trained strategy has never seen these 60 freight missions before.
Table IV shows the performance of three methods on the testing set. 
Compared to Table III, the metrics of both the NC and NE-based methods on the testing set of 60 missions are similar to those in the training set, indicating that the initial value distribution of missions in the testing set is identical to that in the training set. TA-QMIX exhibits performance similar to that in the training set and also achieves a higher fuel saving rate $F_r$, platoon journey rate $P_j$, participation rate $P_r$, and function value $F_v$ compared to the NE-based method in the testing set. This demonstrates that TA-QMIX can generalize to freight tasks with similar distributions.

Next, Fig. \ref{figevaluate} presents a detailed comparison of six metrics between TA-QMIX and the NE-based method across 60 freight missions. Since the NE-based method strictly uses the remaining delay time to plan departure times, resulting in no trucks potentially exceeding time limits, we do not compare the timeout rate $T_o$ between the two methods. 
We use gray to represent the performance of the NE-based method, while other colors represent the performance of TA-QMIX under each metric.
As shown in Fig. \ref{figevaluatea}, \ref{figevaluateb}, and \ref{figevaluated}, the distribution of fuel saving rate, platoon journey rate, and participation rate for the NE-based method across 60 freight missions are similar, indicating that the fuel savings for the NE-based method primarily come from forming platoons.
In Fig. \ref{figevaluatee} and \ref{figevaluatef}, the NE-based method exhibits low time delays, and the distribution is also similar to that of the platoon journey rate, confirming that the NE-based method coordinates platoons by increasing waiting times.
On the other hand, the distributions of fuel saving rate and platoon journey rate for TA-QMIX are not similar, suggesting that trucks save additional fuel by adjusting speeds, in addition to forming platoons. Referring to Fig. \ref{figevaluated}, TA-QMIX achieves higher function values for each mission compared to the NE-based method, indicating that TA-QMIX better balances the trade-off between fuel and time. This further validates the advantage of TA-QMIX in forming platoons through speed adjustments.

\subsubsection{Ablation Experiment}
In TA-QMIX, we introduced TCA Block and TSA Block to enhance the network's representational and training effectiveness. Subsequently, we conducted an ablation experiment to discuss the effectiveness of the TCA Block and TSA Block. 
We used the same training parameters and testing methods to train QMIX+TCA and QMIX+TSA.
Table IV presents the metrics for the four methods.

\begin{table*}[t]
\label{comparetable}
\caption{Ablation Experimental Results in 60 Freight Missions.}
\centering
\resizebox{0.99\textwidth}{!}{
    \begin{tabular}{cccccccc}
    
    \toprule
    Coordination Method & $F_r[\%] \uparrow $     & $T_d[min]\downarrow$  &$T_r[\%]\downarrow$   & $P_j[\%] \uparrow$     & $F_v \uparrow$     & $T_o[\%]\downarrow$     & $P_r[\%] \uparrow$     \\
    \cmidrule(r){1-8}
    QMIX                 & $11.00 \pm 0.89$ & $8.64 \pm 1.79$ & $4.80 \pm 0.69$ & $24.44 \pm 6.17$ & $3966.23 \pm 1491.29$ & $11.42 \pm 4.00$ & $35.73 \pm 8.14$\\
    QMIX+TCA             & $12.45 \pm 1.75$ & \boldmath{$8.42 \pm 2.46$} & \boldmath{$4.68 \pm 1.17$} & $26.27 \pm 5.25$ & $4863.85 \pm 1402.13$ & $1.98 \pm 1.02$ & $44.52 \pm 8.44$\\
    QMIX+TSA               & \boldmath{$15.62 \pm 0.93$} & $14.99 \pm 2.89$ & $8.29 \pm 0.64$ & $28.14 \pm 7.58$ & $3847.74 \pm 1799.25$ & $26.72 \pm 6.30$ & \boldmath{$52.09 \pm 7.98$}\\
    TA-QMIX             &  $13.56 \pm 1.14$ & $10.16 \pm 2.42$ & $5.58 \pm 0.61$ & \boldmath{$30.53 \pm 6.76$} & \boldmath{$5962.79 \pm 1765.53$} & \boldmath{$0.22 \pm 0.45$} & $49.39 \pm 10.56$\\ \bottomrule
    \end{tabular}}
\end{table*}

\begin{table*}[t]
\label{largetable}
\caption{Comparative Experimental Results in Six Large-scale Freight Missions.}
\centering
\resizebox{0.99\textwidth}{!}{
    \begin{tabular}{ccccccccc}
    
    \toprule

    $|\mathcal{N}|$ & Coordination Method & $F_r[\%] \uparrow $     & $T_d[min]\downarrow$  &$T_r[\%]\downarrow$   & $P_j[\%] \uparrow$     & $F_v \uparrow$     & $T_o[\%]\downarrow$     & $P_r[\%] \uparrow$     \\
    \hline
    \multirow{3}{*}{500}                  
                & NC & $1.13 \pm 0.08$ & $-$ & $-$ & $4.15 \pm 0.49$ & $931.72 \pm 96.69$ & $-$ & $8.72 \pm 0.88$\\
                & NE-based       & $6.92$ & \boldmath{$3.45$} & \boldmath{$0.82$} & \boldmath{$39.19$} & $7193.45$ & $-$ & $62.40$\\
                & TA-QMIX & \boldmath{$16.64 \pm 0.29$} & $23.76 \pm 0.79$ & $5.57 \pm 0.11$ & $38.86 \pm 0.91$ & \boldmath{$8835.85 \pm 209.27$} & $0.52 \pm 0.24$ & \boldmath{$68.80 \pm 1.97$}\\
    \hline
    \multirow{3}{*}{1000}                  
                & NC & $1.51 \pm 0.03$ & $-$ & $-$ & $6.43 \pm 0.20$ & $1413.40 \pm 40.35$ & $-$ & $15.08 \pm 0.72$\\
                & NE-based      & $9.97$ & \boldmath{$4.62$} & \boldmath{$1.10$} & \boldmath{$51.36$} & $10207.94$ & $-$ & $77.40$\\
                & TA-QMIX & \boldmath{$17.49 \pm 0.17$} & $20.00 \pm 0.49$ & $4.69 \pm 0.13$ & $48.86 \pm 0.46$ & \boldmath{$11278.32 \pm 85.24$} & $0.74 \pm 0.14$ & \boldmath{$79.58 \pm 1.12$}\\
    \hline
    \multirow{3}{*}{2000}                  
                & NC & $3.10 \pm 0.04$ & $-$ & $-$ & $15.67 \pm 0.22$ & $3427.97 \pm 48.49$ & $-$ & $29.89 \pm 0.40$\\
                & NE-based  & $12.34$ & \boldmath{$4.79$} & \boldmath{$1.14$} & \boldmath{$59.78$} & $13474.13$ & $-$ & $84.90$\\
                & TA-QMIX  & \boldmath{$18.02 \pm 0.18$} & $15.21 \pm 0.43$ & $3.58 \pm 0.13$ & $59.41 \pm 0.92$ & \boldmath{$14086.92 \pm 236.90$} & $0.32 \pm 0.17$ & \boldmath{$87.41 \pm 0.56$}\\
    \hline
    \multirow{3}{*}{3000}                  
                & NC & $3.83 \pm 0.05$ & $-$ & $-$ & $19.82 \pm 0.25$ & $4349.39 \pm 60.85$ & $-$ & $38.23 \pm 0.31$\\
                & NE-based       & $14.32$ & \boldmath{$5.13$} & \boldmath{$1.23$} & \boldmath{$65.78$} & \boldmath{$15818.49$} & $-$ & $89.03$\\
                & TA-QMIX & \boldmath{$18.53 \pm 0.12$} & $12.85 \pm 0.13$ & $3.03 \pm 0.03$ & $64.99 \pm 0.49$ & $15710.17 \pm 137.91$ & $0.30 \pm 0.12$ & \boldmath{$90.42 \pm 0.77$}\\
    \hline
    \multirow{3}{*}{4000}                  
                & NC & $4.94 \pm 0.05$ & $-$ & $-$ & $25.81 \pm 0.32$ & $5762.71 \pm 67.03$ & $-$ & $46.19 \pm 0.48$\\
                & NE-based      & $15.63$ & \boldmath{$5.19$} & \boldmath{$1.23$} & \boldmath{$69.71$} & \boldmath{$17471.79$} & $-$ & $91.78$\\
                & TA-QMIX & \boldmath{$18.95 \pm 0.10$} & $11.01 \pm 0.26$ & $2.61 \pm 0.07$ & $69.12 \pm 0.46$ & $16970.62 \pm 145.69$ & $0.23 \pm 0.07$ & \boldmath{$92.80 \pm 0.10$}\\
    \hline
    \multirow{3}{*}{5000}                  
                & NC & $5.70 \pm 0.02$ & $-$ & $-$ & $29.45 \pm 0.18$ & $6723.34 \pm 29.37$ & $-$ & $51.85 \pm 0.33$\\
                & NE-based       & $16.54$ & \boldmath{$5.041$} & \boldmath{$1.20$} & \boldmath{$72.49$} & \boldmath{$18686.94$} & $-$ & $93.12$\\
                & TA-QMIX & \boldmath{$19.17 \pm 0.09$} & $9.57 \pm 0.16$ & $2.25 \pm 0.02$ & $72.34 \pm 0.21$ & $17990.52 \pm 15.39$ & $0.15 \pm 0.07$ & \boldmath{$94.39 \pm 0.23$}\\
    \bottomrule

    \end{tabular}}
    
\end{table*}

Compared to QMIX, all metrics of QMIX+TCA have improved, significantly reducing the timeout rate. Notably, QMIX+TCA has the lowest delay time and delay time rate among all methods, indicating that the TCA block effectively balances time delays while promoting truck cooperation by processing time and truck information in discrete chunks.
QMIX+TSA achieves the highest fuel-saving rate and participation rate but introduces significant delays, resulting in a lower function value. 
The average delay time rate of 8.29\% is close to the bound of 10\%, and the timeout rate reached 26.72\%, indicating that while the TSA block can enhance cooperation efficiency, it lacks a balance with time efficiency.
Compared to QMIX+TCA, TA-QMIX increased some delay time to adjust speeds for forming platoons, resulting in higher fuel savings and cooperation rates. In comparison to QMIX+TSA, although TA-QMIX slightly reduced the fuel savings rate, it significantly lowered the timeout rate. TA-QMIX achieved the highest function value and the platoon journey rate, indicating that by integrating both TCA block and TSA block, TA-QMIX can further enhance truck cooperation while maintaining a balance with time management.

\subsubsection{Performance in Large-scale Transportation Networks}
In this section, we verify the performance of our method in large-scale transportation networks with total truck counts of $500$, $1000$, $2000$, $3000$, $4000$, and $5000$.
We initialize a freight mission in the total transportation network for each quantity separately. Due to the random errors, we repeat each experiment 5 times for each quantity. Specifically, the NE method does not consider random errors.

As the number of trucks increases, the traffic flow density in the network also increases. Even without using any coordination strategy, there is a probability that trucks will depart from the same hub at the same time, thereby gaining profits. However, after adopting a coordinated strategy, all indicators have greatly improved. The advantage of the NE-based method is that the delay time $T_d$ and the delay time rate $T_r$ are lower, and the truck can obtain fuel benefits while saving time. The advantage of the TA-QMIX method is that it significantly reduces fuel consumption (i.e. $F_r$ is higher) while maintaining a delay time within a tolerable range (i.e. $T_r<(\mathcal{K}_d - 1)$). Through TA-QMIX, trucks can adjust their speed appropriately during the journey, which can reduce fuel consumption while obtaining more opportunities to form platoons (i.e. $P_r$ is higher).

On the other hand, when $\mathcal{N}=500,1000,2000$, the function value $F_v$ of the TA-QMIX method is higher than that of the NE-based method. This indicates that in sparse traffic, the coordination method of waiting only at hubs has fewer opportunities to form platoons, resulting in less fuel savings. By adjusting the speed and directly forming a platoon on the edge, there are more opportunities to form a platoon, which can save more fuel while obtaining a similar platoon journey rate $P_r$.
As the traffic flow density increases ($\mathcal{N}=3000,4000,5000$), the function value $F_v$ of the NE-based method is higher. Due to more opportunities to form platoons, the fuel-saving rate $F_r$ of the two methods is close. Lower time delay cost brings to higher function values. The above analysis indicates that the trained policy can adapt to various traffic flow density situations and large-scale transportation networks. It achieves performance similar to NE-based methods using centralized algorithms and global information calculation in various transportation scenarios.
Furthermore, TA-QMIX exhibits strong robustness as its testing environment accounts for random errors, whereas the NE-based method does not consider such errors in its testing environments.


Next, we selected the experiment with 5000 trucks to compare the computational time of the two methods.
Fig. \ref{figtime} illustrates the relationship between the number of trucks in the current transportation network and the decision-making time for the trucks. Since the NE-based method is a centralized algorithm, its decision-making time increases with the number of trucks. When the number of trucks approaches 5000, the decision time per iteration is close to 500 seconds. This presents a significant obstacle for centralized algorithms when coordinating truck platooning in large-scale transportation networks. In contrast, our proposed method has the capability of distributed execution and is not affected by the number of trucks, with each truck's decision time requiring only 0.001 seconds.

\begin{figure}[t]
	\centering
        \includegraphics[scale=0.48]{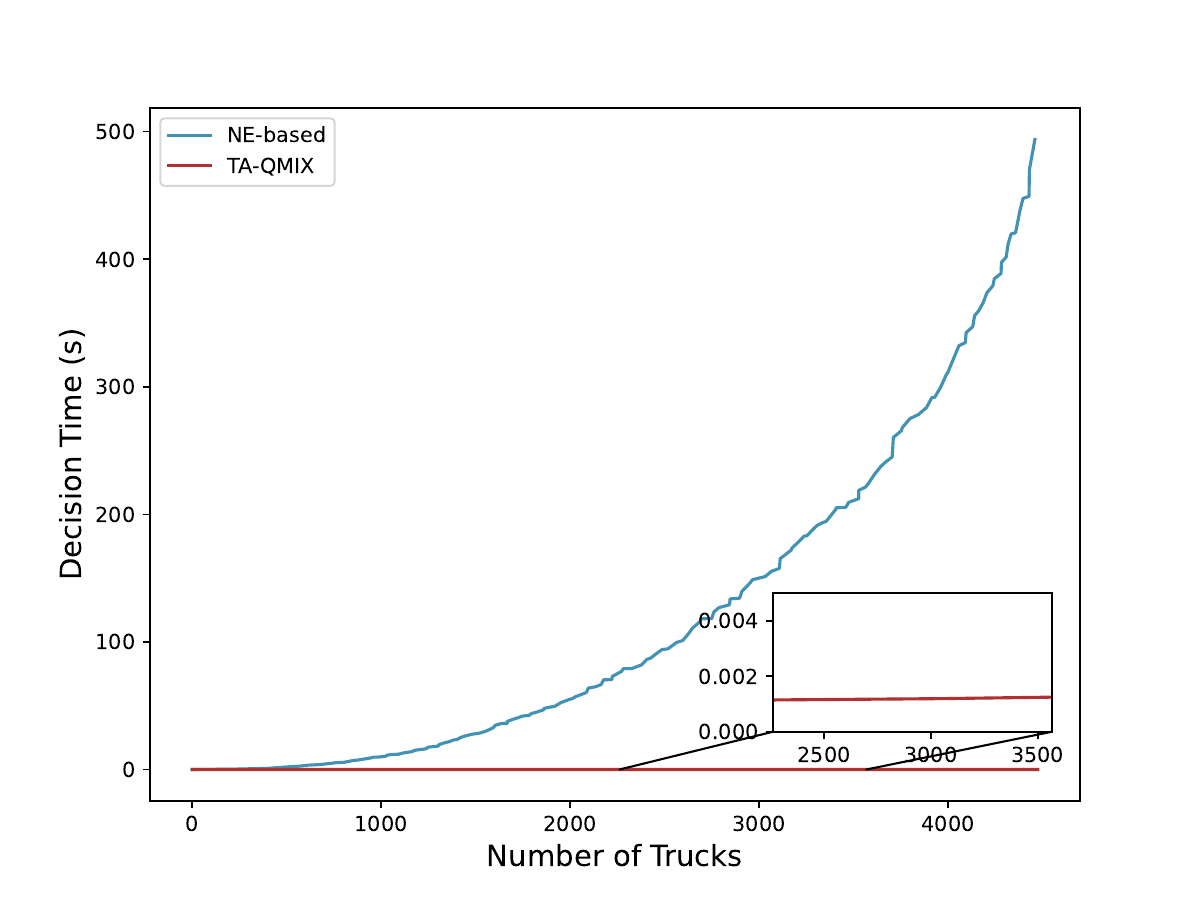}
	\caption{computation time in the large transportation network with 5,000 trucks.}
	\label{figtime}
\end{figure}

\section{Conclusion}
In this paper, we address the joint decision-making process for truck speed and departure time at hubs within the context of platoon coordination in large-scale transportation networks. This problem is a complex integer programming challenge within a dynamic and stochastic transportation environment.
We propose a distributed communication framework that equips trucks with local information for informed decision-making. The problem is reformulated as a Dec-POMDP, with well-defined state transitions and dense rewards to facilitate training.
Furthermore, we apply the TA-QMIX deep reinforcement learning framework to optimize speed and time planning for platoon coordination in extensive transportation networks. TA-QMIX leverages an attention mechanism to enhance the representation of environmental states and truck-specific information, thereby improving cooperative behavior among trucks in the platoon coordination problem.
To validate our approach, we developed a transportation network simulator for the Yangtze River Delta region in China. Simulation experiments demonstrate the effectiveness and generalization of our proposed method.
Our future work will focus on exploring dynamic path planning for trucks during transit, aiming to enrich the decision-making space and improve collaboration potential among trucks.


\section*{Acknowledgments}



%

\bibliographystyle{IEEEtran}
\small
\bibliography{refer}

\end{document}